\documentclass[review]{elsarticle}
\usepackage{lineno,hyperref}
\modulolinenumbers[5]

\journal{Physics of Fluids}

\usepackage{amsmath}
\usepackage{dsfont}
\usepackage{amssymb}
\usepackage{graphicx,color}
\usepackage{extarrows}
\usepackage{mathrsfs}
\usepackage{booktabs}
\usepackage{multirow}
\usepackage{threeparttable}
\usepackage[justification=raggedright]{caption}
\usepackage{threeparttable}
\usepackage[misc]{ifsym}
\usepackage{array}
\usepackage{graphicx}
\usepackage{float}
\usepackage{subfigure}

\newtheorem{thm}{Theorem}[section]

\makeatletter
\@addtoreset{equation}{section}
\makeatother

\graphicspath{ {./images/} }
\DeclareMathOperator*{\argmin}{arg\,min}

\begin{document}
	\begin{frontmatter}
		
		\title{\bf\Large Data-driven Parameter Inference for Nonlinear Wave Equations with Markovian Switching}
		
		\author{Yi Zhang \fnref{addr1}}
		\author{Zhikun Zhang \fnref{addr1}\corref{mycorrespondingauthor}}
		\ead{zkzhang@hust.edu.cn}
		\author{Xiangjun Wang \fnref{addr1}}
		\cortext[mycorrespondingauthor]{Corresponding author}
		\ead{xjwang@hust.edu.cn}
		
		\address[addr1]{School of Mathematics and Statistics,\\
			Huazhong University of Science and Technology, Wuhan 430074, China}

		\begin{abstract}
			Traditional partial differential equations with constant coefficients often struggle to capture abrupt changes in real-world phenomena, leading to the development of variable coefficient PDEs and Markovian switching models. Recently, research has introduced the concept of PDEs with Markov switching models, established their well-posedness, and presented numerical methods. However, there has been limited discussion on parameter estimation for the jump coefficients in these models. This paper addresses this gap by focusing on parameter inference for the wave equation with Markovian switching. We propose a Bayesian statistical framework using discrete sparse Bayesian learning to establish its convergence and a uniform error bound. Our method requires fewer assumptions and enables independent parameter inference for each segment by allowing different underlying structures for the parameter estimation problem within each segmented time interval. The effectiveness of our approach is demonstrated through three numerical cases, which involve noisy spatiotemporal data from different wave equations with Markovian switching. The results show strong performance in parameter estimation for variable coefficient PDEs.
			
		\end{abstract}
		
		\begin{keyword}
			Wave equation; Markov switching model; Parameter inference; Inverse problem; Bayesian statistics
		\end{keyword}
		
	\end{frontmatter}

	\section{Introduction}
	Partial differential equations(PDEs) are crucial in various fields like image processing\cite{Reuter2024pof,Halim2020,LOTFI2022}, medical imaging\cite{Mang2018}, and machine learning\cite{FANG201714,BURGER2021, NIPS2021} for their ability to model real-world phenomena. In the automatic control field, they describe state changes over time and space, aiding in stability and control design\cite{Wang2023,DARBHA1999}. PDEs also model fault monitoring in distributed systems\cite{ferdowsi2019} and simulate chemical diffusion in biological cells\cite{A.Talkington2015}. However, most of these systems modeled by fixed coefficient equations often exhibit jumps due to perturbations from environmental changes or interventions, and such perturbations can abruptly alter system structures or parameters\cite{Ren2020pof,Shi2015}. 
	
	To better describe system changes, variable coefficient PDEs have been proposed with theoretical and practical advancements. For instance, Muhammad Sohail and S. Abbas examined the flow of Prandtl fluid influenced by variable coefficient PDEs, focusing on heat and mass transfer using advanced diffusion and conductivity models\cite{Sohail2024}. Metcalfe and Tataru constructed outgoing parametrices for time-dependent wave equations\cite{Metcalfe2012}. Yin \textit{et.al.} derived a variable coefficient $(2+1)$-dimensional extended cKP equation for Rossby waves, introducing a new transformation to obtain exact solutions\cite{Yin2023pof}. Ning et al. examined the stability of wave equations with variable coefficients and nonlinear boundary feedback\cite{NING2014}. Compared to variable coefficient PDEs, the Markovian switching model introduced by Krasovskii in 1961\cite{Krasovskii1961} is more adept at handling abrupt changes in system behavior. Early studies about Markovian switching focused on controllability and stability. Practical applications emerged with Sworder's work on hybrid linear systems affected by Markov jump parameters, providing optimal control feedback\cite{Sworder1969}. Markovian switching systems have since been applied in various fields. A notable instance is economics, where Hamilton utilized a Markovian switching model to represent time series behaviors across different regimes. By allowing transitions between multiple structures, this model effectively captures more complex dynamic patterns\cite{HAMILTON1988, HAMILTON1989}.
	
	Even though both variable coefficient PDEs and the Markovian switching model can address the jumps in a system's structure or parameters, they still face challenges in capturing the dynamic changes in space. Therefore, Li and Wang \textit{et al.}\cite{J.Li2023} recently proposed a model called diffusion equation with Markovian switching. This model integrates the spatial descriptive power of PDEs with the stochastic nature of Markovian switching mechanisms, allowing for the simultaneous consideration of spatial distribution and temporal jumps in the system. They provided proof of well-posedness, stability, and numerical generation methods using non-uniform grids. Diffusion equations effectively model processes involving gradual changes such as heat conduction, while wave equations are suited for dynamic phenomena like sound and electromagnetic waves. Incorporating Markovian switching allows wave equations to account for abrupt changes, expanding their applicability. Thus They also constructed the well-posedness for the wave equation with Markovian switching\cite{J.Li2024}. On the other hand, the inverse problems of some models mentioned above, \textit{i.e.} inferring parameters from observed data, have garnered attention. Yuan Ye \textit{et.al.} proposed a framework discovering underlying PDEs and their parameters by sparse Bayesian learning(SBL)\cite{S3d}. D.A. Barajas-Solano presents two approximate Bayesian inference methods for estimating parameters in PDE models with space-dependent and state-dependent parameters, focusing on diffusion coefficients in linear and nonlinear diffusion equations\cite{BARA2019}. Lu H \textit{et.al.} introduced fully nonlinear free surface Physics-Informed Neural Networks to address nonlinear free surface wave propagation and inverse problems in wave modeling\cite{Lu2024pof}. 
	Xu \textit{et.al.} proposed a spatiotemporal parallel PINNs framework to efficiently solve inverse problems in fluid mechanics for high Reynolds number turbulent flows\cite{xu2023pof}. Li \textit{et.al.} estimated parameters of a one-dimensional reaction-diffusion model named Fisher’s equation with Markovian switching where the Markov process represents the diffusion parameter\cite{J.Li2023}.
	
	In our paper, we further advance the Discrete Sparse Bayesian Learning(DSBL) algorithm specifically tailored for the wave equation with Markovian switching, addressing the inverse problem of inferring jump parameters. We conduct a comprehensive convergence analysis and establish a uniform error bound for the proposed method, ensuring its robustness and reliability. To validate the effectiveness of our approach, we design a series of numerical experiments utilizing spatiotemporal data from different nonlinear wave equations, including the Sine-Gordon equation, the Klein-Gordon equation, and a two-dimensional nonhomogeneous wave equation with Markovian switching. These experiments demonstrate the DSBL method's accuracy in estimating the parameters of wave equations with Markovian switching, as well as its strong adaptability to the dynamic switching of system coefficients. The results highlight the method's potential for effectively handling complex inverse problems in systems with variable coefficients and stochastic switching behavior.
	
	The paper is structured as follows: Section 2 introduces the definition of the wave equation with Markovian switching. Subsequently, the DSBL algorithm is developed for this system, with a detailed discussion of its convergence property and the error bound. In Section 3, we generate spatiotemporal data through numerical methods and conduct three numerical experiments of increasing complexity, which demonstrate the algorithm's accuracy in parameter estimation. Our method significantly outperforms a single model without Markovian switching. Finally, Section 4 offers conclusions and prospects for future work.
	
	\section{Wave equation with Markovian switching}
	\subsection{Definition of Wave equation with Markovian switching}
	In the real world, the parameters of complex systems usually vary over time. Such coefficient changes can follow stochastic processes including the Markov process, denoted by $M_t$. In this paper, we characterize wave equation with Markovian switching in the following form
	\begin{equation}\label{Wave system with Mt}
		\begin{cases}
			u_{tt} - M_t\Delta u  = f(u), & \textit{in}\  U\times (0,T], \\
			u(x,t) = 0 ,& \textit{on} \ \partial U\times [0,T], \\
			u(x,0) = g(x), &  \textit{on}\ U\times \{t=0\}, 
		\end{cases}
	\end{equation}
	where $g(x)$ is the initial value, and $f(u)$ is a nonlinear function. The global time $T$ is fixed and $U \subseteq \mathbb{R}$ is open and bounded. A stochastic process $M_t=\{M(t)|\:0\leq t \leq T\}$ is a Markov process with a state space $\mathcal{E}$. For any $0\leq t_1<t_2<...<t_n\leq T$, there are corresponding states $m_1,m_2,...,m_n\in \mathcal{E}$. Therefore, $M_t$ represents the parameters of the complex system in each state. In particular, the Markovian property regulates that the current value of the state variable depends on its immediate past value. When the system transitions to a new state, the initial value in the current state is the value taken at the end of the previous state, ensuring the continuity of the system.
	
	\subsection{Parameter inference for the wave equation with Markovian switching}
	Given the structure of PDEs, the parameter inference problem can be converted to a regression problem and then solved using optimization algorithms. Considering the wave systems with Markovian switching, the general form can be rewritten as
	\begin{equation}\label{general_type}
		\frac{\partial^2 u(x,t)}{\partial t^2}= G\left(x, u(x,t), \frac{\partial u(x,t)}{\partial x_1}, \ldots, \frac{\partial u(x,t)}{\partial x_n}, \frac{\partial^2 u(x,t)}{\partial x_1^2}, \ldots, \frac{\partial^2 u(x,t)}{\partial x_1 \partial x_n}, \ldots, M_t; \Theta\right),
	\end{equation}
	where $t \in [0, +\infty)$, $x=(x_1,...,x_n)^T \in \mathbb{R}^n$, $G$ is an unknown function that depends on the dynamical variable $u$ and its derivatives, $M_t$ is a Markov chain, and $\Theta$ represents a parameter switching space. Note that the Markov chain $M_t$ governs the parameter vector set $\vartheta=\{\theta_1,\theta_2,...,\theta_d\}$, which randomly varies within space $\Theta$. Our aim is to learn $G$ and $\vartheta$, thus given a instantaneously varying parameters vector $\theta \in \vartheta$, the Eq.\ref{general_type} can be written as the following regression form
	\begin{equation}\label{Regression}
		u_{tt}(\textbf{x},\textbf{t}) = D(t)\cdot \theta+\mathbf{\varepsilon}.
	\end{equation}
	Alternatively, by substituting spatiotemporal data into Eq.\ref{Regression}, we obtain
	\begin{equation}\label{Regression_trans}
		\underbrace{
			\begin{bmatrix}
				u_{tt}(x_1, t_1) \\
				u_{tt}(x_2, t_1) \\
				\vdots \\
				u_{tt}(x_{n}, t_{m})
			\end{bmatrix}
		}_{\mathbf{y}}
		=
		\underbrace{
			\begin{bmatrix}
				1 & u(x_1, t_1) & u_x(x_1, t_1) & \cdots & u^2 u_{xx}(x_1, t_1) \\
				1 & u(x_2, t_1) & u_x(x_2, t_1) & \cdots & u^2 u_{xx}(x_2, t_1) \\
				\vdots & \vdots & \vdots & \ddots & \vdots \\
				1 & u(x_{n}, t_{m}) & u_x(x_{n}, t_{n}) & \cdots & u^2 u_{xx}(x_{n}, t_{m})
			\end{bmatrix}
		}_{\text{Dictionary matrix } \mathbf{D}}
		\underbrace{
			\begin{bmatrix}
				\theta_1 \\
				\theta_2 \\
				\vdots \\
				\theta_m
			\end{bmatrix}
		}_{\mathbf{\theta}}
		+
		\underbrace{
			\begin{bmatrix}
				\varepsilon_1 \\
				\varepsilon_2 \\
				\vdots \\
				\varepsilon_m
			\end{bmatrix}
		}_{\mathbf{\varepsilon}}
	\end{equation}
	
	In Eq.\ref{Regression}, \(D \in \mathbb{R}^{n \times m}\) is a dictionary matrix composed of various combinations of terms (e.g., \(u\), \(u_x\), \(u_{xx}\), which can be combined as \(uu_x\), \(uu_{xx}\), \(u_x u_{xx}\)) related to the structure of the PDEs. $\varepsilon\sim \mathcal{N}(0,\sigma^2I_n)$ is an \textit{i.i.d.} noise vector.
	
	Therefore, various methods can be applied to Eq.\ref{Regression}. In our paper, we primarily develop the SBL method into a discrete optimization algorithm called DSBL. For each discrete state, the SBL optimization algorithm is used for parameter estimation. The DSBL method retains the ability to determine the model complexity automatically and encourages sparser coefficients, achieving dimensionality reduction by introducing a sparse prior distribution $p(\theta;\gamma)$ and $\gamma$ relates to the variance of each component of $\theta$. Similar to SBL, a difference of convex function concerning the parameters $\theta$ and $\gamma$ is derived, serving as the loss function $\mathcal{L}(\theta, \gamma)$. Subsequently, an iterative algorithm, the convex-concave rocedure(CCP) condition, is utilized to solve the following sequence of convex programs
	\begin{equation}\label{CCP}
		(\theta^{(k+1)}, \gamma^{(k+1)}) \in \argmin_{(\theta, \gamma) \in \Omega} \hat{\mathcal{L}}(\theta, \gamma; \theta^{(k)}, \gamma^{(k)}),
	\end{equation}
	where $\Omega = \mathbb{R}^m \times [0, +\infty)$. 
	
	\subsection{Convergence and error bound of DSBL using in the wave systems with Markovian switching}
	
	After presenting the model and its corresponding algorithm, the next essential step is to address the convergence and error bound of the DSBL algorithm. Before that, proving the convergence requires establishing the boundedness of the loss function's and parameters' iterative sequence, which ensures the existence of a convergent subsequence for $(\theta^{(k+1)}, \gamma^{(k+1)})$.
	
	\begin{thm}\label{bounded sequence}
		$\{\mathcal{L}(\theta^{(k)}, \gamma^{(k)})\}_{k=0}^{\infty}$ is a bounded, monotone non-increasing sequence.
	\end{thm}
	
	Once the desired iteration accuracy is achieved, the parameters $\theta^{(k+1)}$ obtained by $(\theta^{(k+1)}, \gamma^{(k+1)})$ at this time represent the wave equation's coefficients, with $M_{t_i}$ determined via the SBL algorithm. Next, the convergence theorem for the DSBL method is presented.
	
	\begin{thm}\label{convergence}
		Assume that in Eq.\ref{Wave system with Mt}, $M_t$ is a continuous, irreducible, aperiodic, and recurrent Markov chain with a finite state space. The convergence of DSBL can be converted to SBL on the local time interval $[\tau_k,\tau_{k+1})$. Then, it holds over \([0, T]\),
		\begin{equation}
			\lim_{k\to \infty} \mathcal{L}(\theta^{(k)},\gamma^{(k)})=\mathcal{L}(\theta^*,\gamma^*),
		\end{equation}
		where $(\theta^*,\gamma^*)$ is a stationary point of $\mathcal{L}(\theta,\gamma)$.
	\end{thm}
	
	As Theorem~\ref{convergence} demonstrates, the iteration sequence of DSBL \(\{(\theta^{(k)},\gamma^{(k)})\}_{k=0}^{\infty}\) converges to a stationary point of \(\mathcal{L}(\theta,\gamma)\). Therefore, it is necessary to explore the relationship between the true value \(\theta^{\text{true}}\) and the estimation \(\theta^*\) of the equation's parameter.
	\begin{thm}\label{Error bound}
		Suppose that system \ref{general_type} involves $d$ parameters and the Markov chain $M_t$ possesses $K$ finite states. Denote \(\theta_{\text{min}}\) as the minimum absolute value of all nonzero elements in \(\theta_{\text{true}}\). If system \ref{general_type} satisfies $D^TD = I_m$, then the error bound is given by
		\begin{equation}
			|\theta^{true} - \theta^*|
			\leq \sqrt{dK}\frac{\theta_{\min}^2-4\sigma\theta_{\min}+8\sigma^2}{2\theta_{\min}-4\sigma}.
		\end{equation}
	\end{thm}
	
	Notice that when $\sigma^2 \rightarrow 0$, the conditions of Theorem~\ref{Error bound} are satisfied. The parameters \(d\) and \(K\) represent the complexity of the model itself and the complexity of the system state, respectively. The error bound of the equation increases with the number of parameters \(d\) and the states \(K\) of the Markov chain. The \(\theta_{\min}\) is typically a relatively small constant, but as \(dK\) becomes larger, the error bound becomes significant due to the diminishing variance \(\sigma\).
	
	The proof of these theorems in this section is given in section \ref{Appendix} as an appendix.
	
	\section{Numerical Experiment}
	In our numerical experiment, we collect spatio-temporal data via the finite difference method and finite element method under non-uniform grids to obtain the numerical solution of partial differential equations with Markovian switching. The data collected is composed into a snapshot matrix and the $ith$ row and $jth$ column of the matrix is denoted by $u(x_i,t_j)$. According to the previous theoretical derivation, the synthetic data can be written as 
	\begin{equation}
		y(x_i,t_j)=u(x_i,t_j)+\xi_{ij},
	\end{equation}
	where $u(x_i,t_j)$ represents element in the dictionary matrix $D$, $\xi_{ij}$ follows a normal distribution $\mathcal{N}(\mu,\sigma_u)$ based on the magnitude of $u(x_i,t_j)$ with mean $\mu=0$. Through the synthetic data, the time derivative on LHS of \ref{Regression_trans} and the spatial derivative involved in the dictionary can be computed by using the finite difference method.

	\subsection{Case 1: Sine-Gordon Equation with Markovian switching}
	The Sine-Gordon(SG) equation is a nonlinear hyperbolic partial differential equation arising in different applications, including propagation of magnetic flux on Josephson junctions and many others discussed in \cite{Scott1973}. It takes the form
	\begin{equation}
		u_{tt}-u_{xx}=-\sin u,
	\end{equation}
	and admits several types of special solutions, such as traveling-wave solutions, soliton solutions, and small amplitude solutions. Our goal is to extract a dynamical regime from the soliton solutions. 
	
	To be concrete, we consider an initial boundary value problem given by
	\begin{equation}{\label{eq:sine-gordon}}
		\begin{cases}
			u_{tt} - M_tu_{xx} + \alpha \sin(\omega u) = 0, & x \in [0, L], \, t > 0, \\
			u(x, 0) = 10 \exp\left(-\left(x - \frac{L}{2}\right)^2\right), & x \in [0, L], \\
			u_t(x, 0) = 0, & x \in [0, L], \\
			u(0, t) = 0, & t > 0,\\
			u(L, t) = 0, & t > 0,
		\end{cases}
	\end{equation}
	where $\alpha$ and $\omega$ represent the parameters of the system. The initial condition $u(x,0)$ is characterized by a Gaussian-shaped curve. It should be noted that $M_t$ represents a continuous  Markov chain generating by a generator matrix $Q_1$ in Eq.\ref{Q1} and taking in the following form as Eq.\ref{CMC}
	\begin{equation}{\label{CMC}}
		M_t = 
		\begin{cases}
			1, & t \in [0.00, 1.22) \cup [3.90, 5.05), \\
			0.5, &  t \in [1.22, 3.23) \cup [5.05, 5.97) \cup [6.67, 8.00], \\
			0.1, & t \in [3.23, 3.90) \cup [5.97, 6.67),
		\end{cases}
	\end{equation}
	and
	\begin{equation}\label{Q1}
		Q_1 = \begin{bmatrix}
			-1.2 & 0.7 & 0.5 \\
			0.3 & -1.0 & 0.7 \\
			0.4 & 0.6 & -1.0
		\end{bmatrix}.
	\end{equation}
	
	The total simulation time of the system is 8s. For our analysis using DSBL, we employ a dataset comprising $400 \times 700$ spatiotemporal data points. This involves uniformly discretizing the spatial domain $[0,10]$, so that L=10, of interest into 400 equal intervals with a grid step of $\Delta x=0.025$. For the temporal domain $[0, 8]$, the simulation time is divided into seven segments based on the states of the Markov chain. Each segment is further subdivided into 100 equal intervals. This segmentation allows the simulation to accurately reflect the varying states of the system over time. To ensure the continuity of the model, the final values of the spatiotemporal data from the previous time segment are used as the initial values for the subsequent time segment when applying the DSBL algorithm.  
	The parameter inference results of the DSBL applied to the SG equation with Markovian switching are summarized in Table \ref{tab:SG}. The visualization of the results is provided in Fig.~\ref{SG_solution}. Furthermore, Fig.~\ref{SG_heat_map3} presents the error heat map, highlighting the discrepancies between numerical and inferred solutions. The results indicate that this method effectively detects system switching and accurately infers the corresponding parameters.
	
	\begin{table}[t]
		\centering
		\begin{tabular}{ccc}
			\toprule
			& Time interval  & SG equation(\textcolor{red}{Absolute error}) \\
			\midrule
			Simulation set & 
			0.00--1.22;3.90--5.05 & $u_{tt} = 1u_{xx}-\sin u$ \\
			& 1.22--3.23;5.05--5.97;6.67--8.00 & $u_{tt} = 0.5u_{xx} -\sin u$ \\
			& 3.23--3.90;5.97--6.67 & $u_{tt} = 0.1u_{xx} -\sin u$ \\
			\midrule
			\textbf{Mixture model} & 0.00--1.22 & $u_{tt} = 0.9994(\textcolor{red}{0.06\%})u_{xx}-1.0004(\textcolor{red}{0.04\%})\sin u$ \\
			& 1.22--3.23 & $u_{tt} = 0.4999(\textcolor{red}{0.01\%})u_{xx}-1.0019(\textcolor{red}{0.19\%})\sin u$ \\
			& 3.23--3.90 & $u_{tt} = 0.0831(\textcolor{red}{1.69\%})u_{xx}-0.9855(\textcolor{red}{1.45\%})\sin u$ \\
			& 3.90--5.05 & $u_{tt} = 1.0003(\textcolor{red}{0.03\%})u_{xx}-1.0023(\textcolor{red}{0.23\%})\sin u$ \\
			& 5.05--5.97 & $u_{tt} = 0.4998(\textcolor{red}{0.02\%})u_{xx}-0.9984(\textcolor{red}{0.16\%})\sin u$ \\
			& 5.97--6.67 & $u_{tt} = 0.0990(\textcolor{red}{0.1\%})u_{xx}-0.9957(\textcolor{red}{0.43\%})\sin u$ \\
			& 6.67--8.00 & $u_{tt} = 0.5001(\textcolor{red}{0.01\%})u_{xx}-1.000(\textcolor{red}{0.00\%})\sin u$ \\
			\bottomrule
		\end{tabular}
		\caption{Parameter inference results of DSBL applied to the SG equation with Markovian Switching.}
		\label{tab:SG}
	\end{table}

	\subsection{Case 2: Klein-Gordon Equation with Markovian switching}
	
	\begin{table}[t]
		\centering
		\begin{tabular}{ccc}
			\toprule
			& Time interval  & KG equation (\textcolor{red}{Absolute error}) \\
			\midrule
			Simulation Set & 0.0--0.45; 2.02--2.5& $u_{tt} = 2u_{xx}-u-u^3$ \\
			& 0.45--1.04; 1.48--2.02 & $u_{tt} = 1u_{xx} -u-u^3 $ \\
			& 1.04--1.48 & $u_{tt} = 0.5u_{xx} -u-u^3$ \\
			\midrule
			\textbf{Mixture Model} & 0.0--0.45 & $u_{tt} = 1.9931(\textcolor{red}{0.69\%})u_{xx}-1.0074(\textcolor{red}{0.74\%})u-1.0000(\textcolor{red}{0.00\%})u^3$ \\
			& 0.45--1.04 & $u_{tt} = 1.0012(\textcolor{red}{0.12\%})u_{xx}-0.9865(\textcolor{red}{1.35\%})u-1.0003(\textcolor{red}{0.03\%})u^3$ \\
			& 1.04--1.48 & $u_{tt} = 0.4980(\textcolor{red}{0.20\%})u_{xx}-1.0382(\textcolor{red}{3.82\%})u-0.9987(\textcolor{red}{0.13\%})u^3$ \\
			& 1.48--2.02 & $u_{tt} = -1.0007(\textcolor{red}{0.07\%})u_{xx}-1.0131(\textcolor{red}{1.31\%})u-0.9820(\textcolor{red}{2.80\%})u^3$ \\
			& 2.0--2.5 & $u_{tt} = 2.0015(\textcolor{red}{0.15\%})u_{xx}-1.0318(\textcolor{red}{2.2\%})u-0.9780(\textcolor{red}{2.20\%})u^3$ \\
			\bottomrule
		\end{tabular}
		\caption{\label{tab:KG}Parameter inference results of DSBL applied to the KG equation with Markovian Switching.}
	\end{table}

	The second case we consider is the Klein-Gordon(KG) equation with a cubic nonlinearity
	\begin{equation}
		u_{tt}=u_{xx}-u-u^3.
	\end{equation}
	The KG equation, related to the Schr\"{o}dinger equation, has many applications. It is used to describe scalar and spinor fields in quantum field theory and relativistic quantum mechanics and finds applications in areas such as spin waves and nonlinear optics\cite{Marx1967, Dodd1982}.
	
	Similarly, with the first case, we generate the data by the finite difference method and consider the following initial boundary value problem
	\begin{equation}{\label{eq:klein-gordon}}
		\begin{cases}
			u_{tt} - M_tu_{xx} = -u-u^3, & x \in [0, L], \, t > 0, \\
			u(x, 0) = 8 \exp\left(-\left(x - \frac{L}{2}\right)^2\right), & x \in [0, L], \\
			u_t(x, 0) = 0, & x \in [0, L], \\
			u(0, t) = 0, & t > 0, \\
			u(L, t) = 0, & t > 0.
		\end{cases}
	\end{equation}
	
	The Markov chain \(M_t\) is discrete in this case as given by Eq.~\ref{DMC}, and its generator matrix is depicted in Eq.~\ref{Q2}.
	\begin{equation}{\label{DMC}}
		M_t = 
		\begin{cases} 
			2, &  t \in [0.00, 0.45) \cup [2.02, 2.5),\\
			1, &  t \in [0.45, 1.04) \cup [1.48, 2.02), \\
			0.5, & t \in [1.04, 1.48),
		\end{cases}
	\end{equation}
	and
	\begin{equation}\label{Q2}
		Q_2 = \begin{bmatrix}
			-1.4 & 0.7 & 0.7 \\
			0.7 & -1.4 & 0.7 \\
			0.7 & 0.7 & -1.4
		\end{bmatrix}.
	\end{equation}
	
	We address the problem Eq.~\ref{eq:klein-gordon} following a similar procedure to that used for the SG equation, except employing different grid partitioning methods. Given the changes in the Markov chain, the simulation time is divided into five segments. Then for each segment, we use a grid size of $512 \times 512$ for space and time.
	
\begin{figure}[p]
		\centering
		\subfigure{
			\centering
			\includegraphics[width=0.65\textwidth]{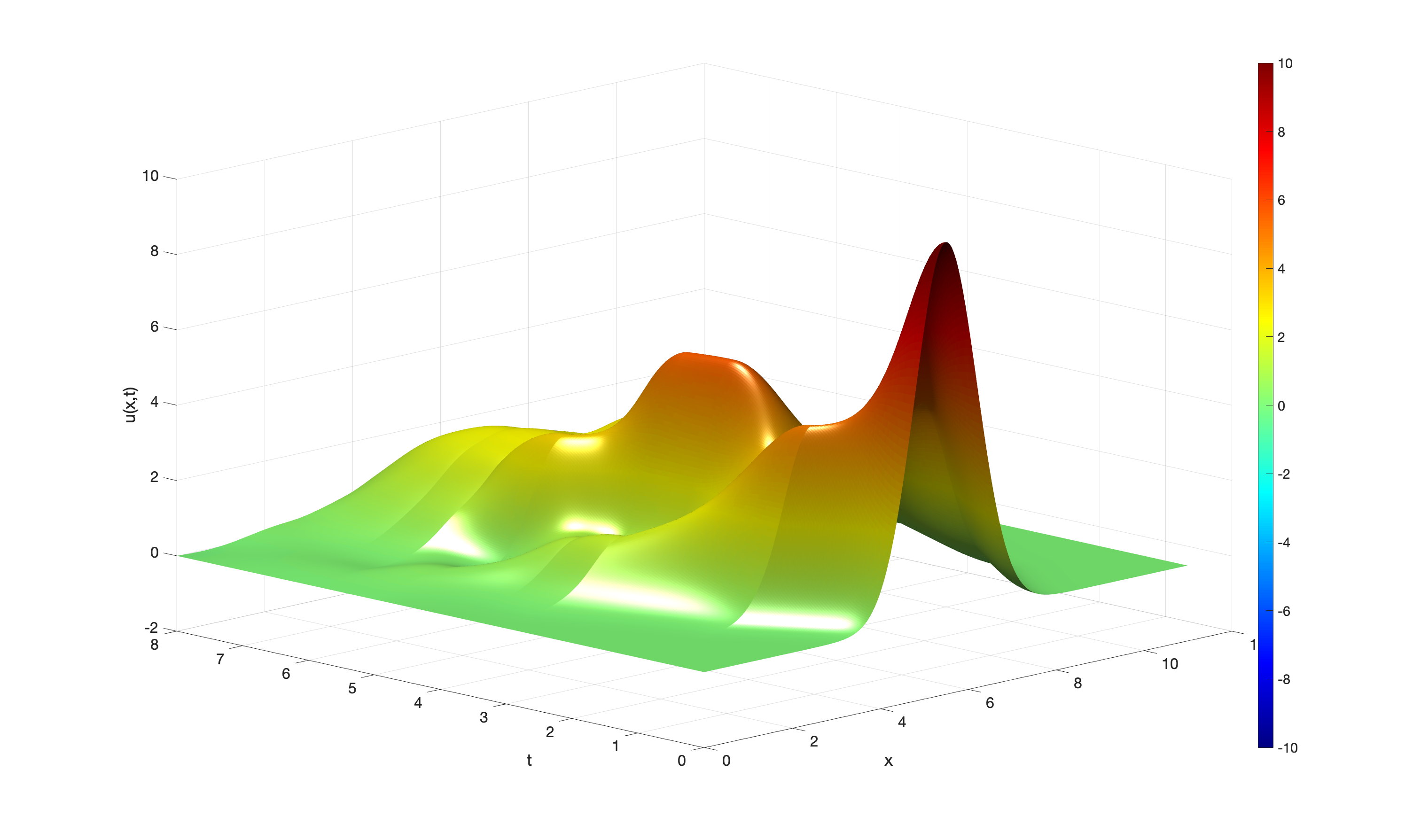}
		}
		\caption{\label{SG_solution} Numerical Solution of SG equation with Markovian Switching}
		\subfigure{
			\centering
			\includegraphics[width=0.65\textwidth]{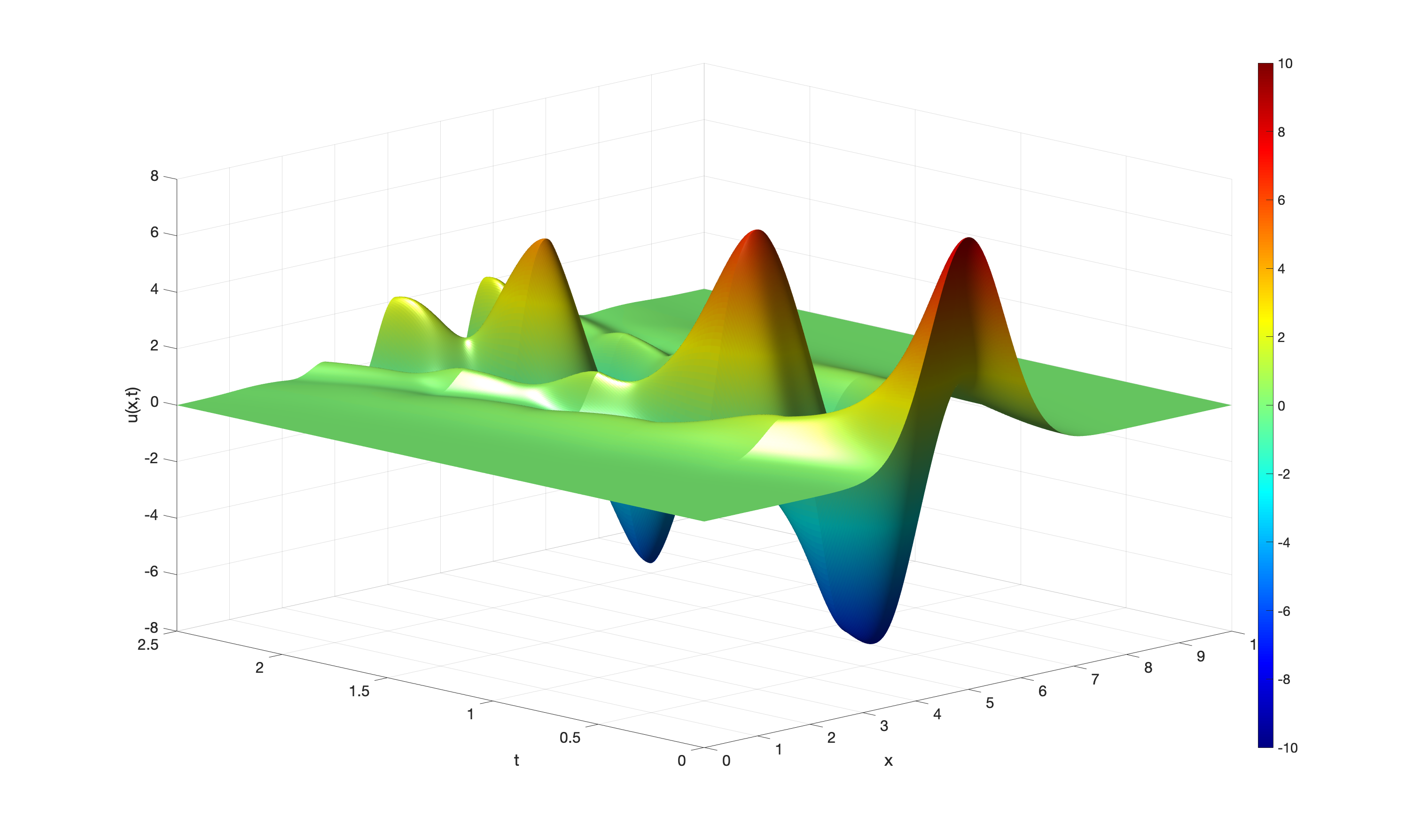}
		}
		\caption{\label{KG_solution} Numerical Solution of KG equation with Markovian Switching}
		\subfigure{
			\centering
			\begin{minipage}{0.325\textwidth}
				\centering
				\includegraphics[width=0.88\textwidth]{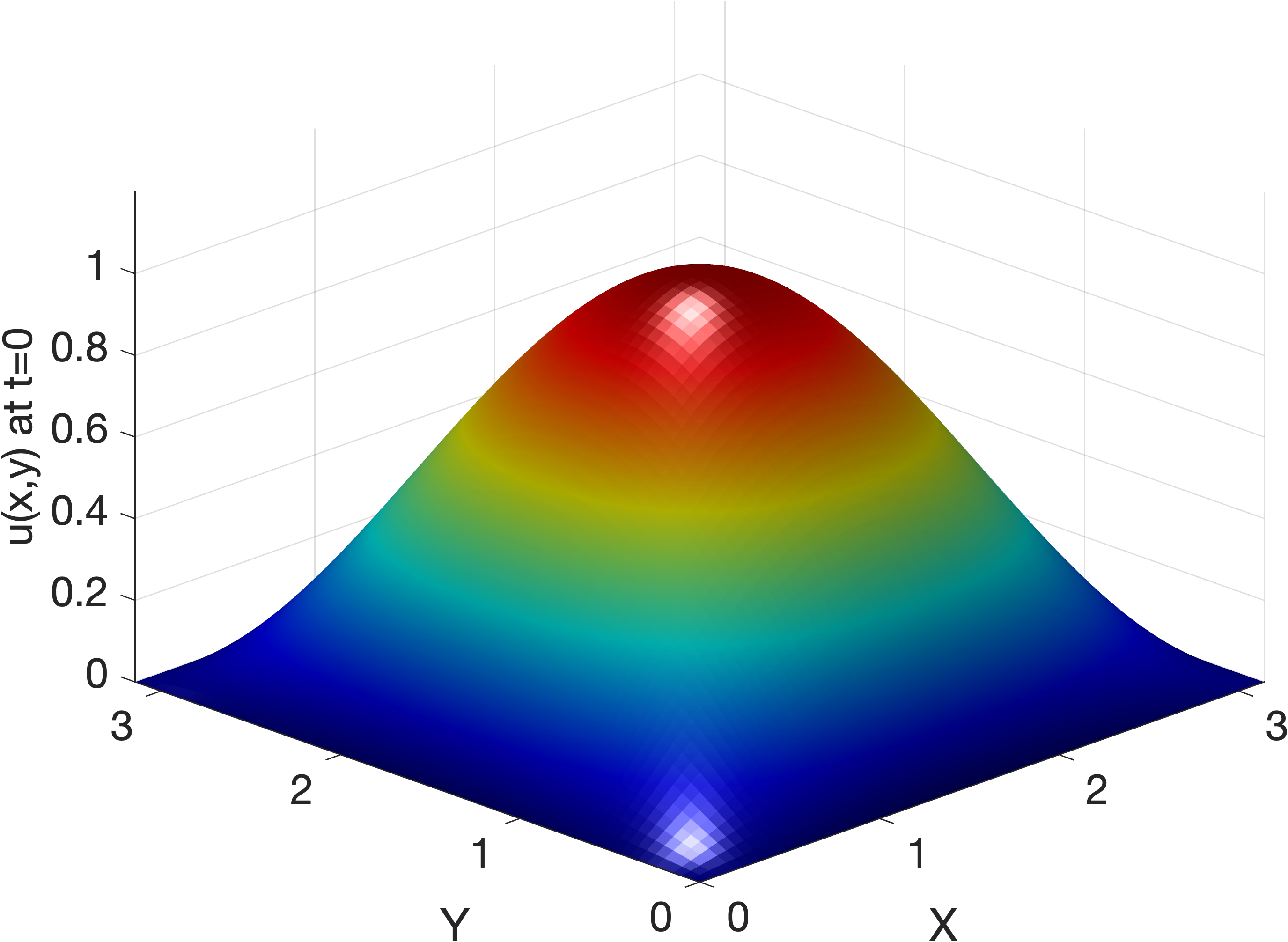}
			\end{minipage}
			\begin{minipage}{0.325\textwidth}
				\centering
				\includegraphics[width=0.88\textwidth]{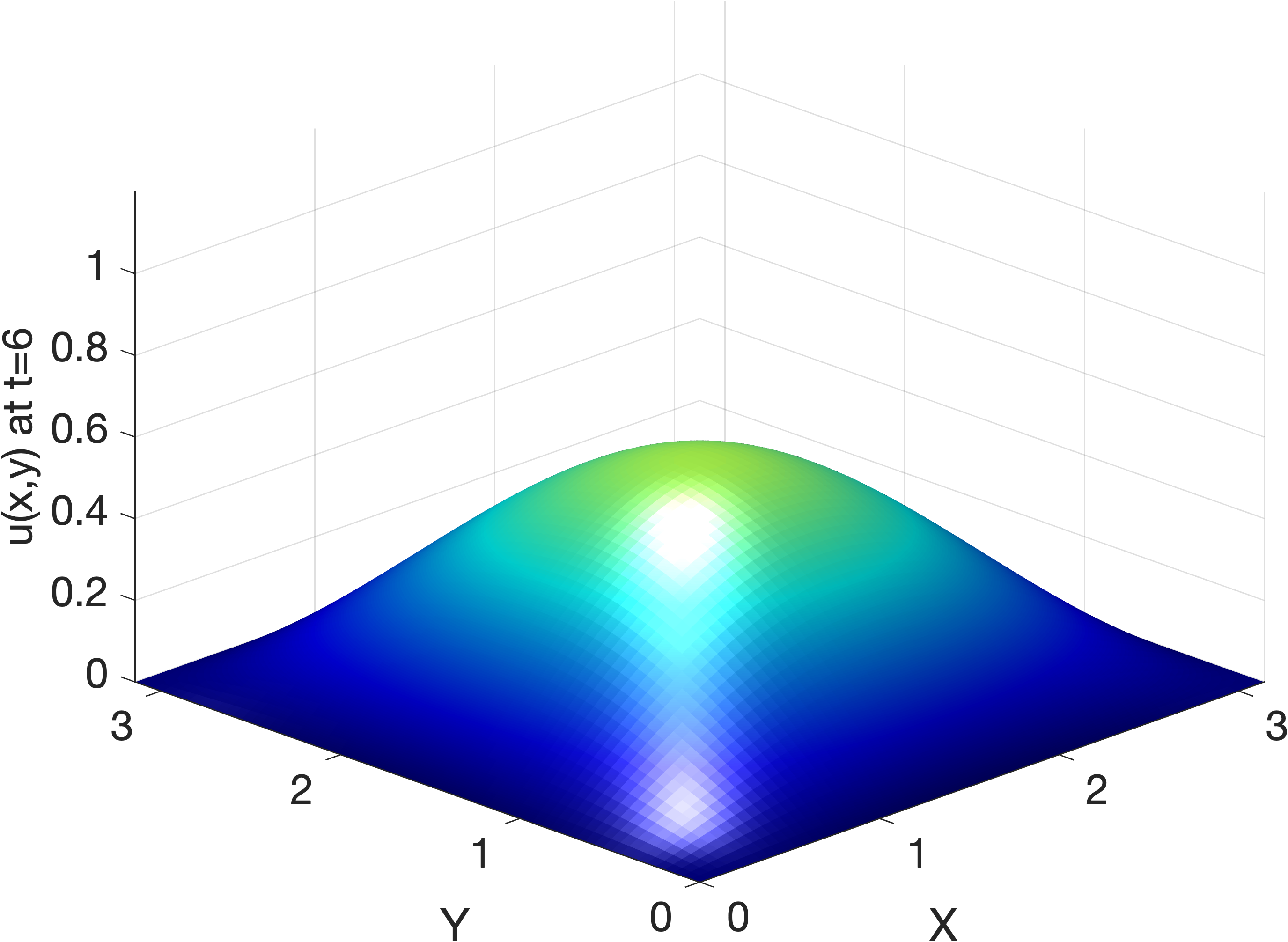}
			\end{minipage}
			\begin{minipage}{0.325\textwidth}
				\centering
				\includegraphics[width=0.88\textwidth]{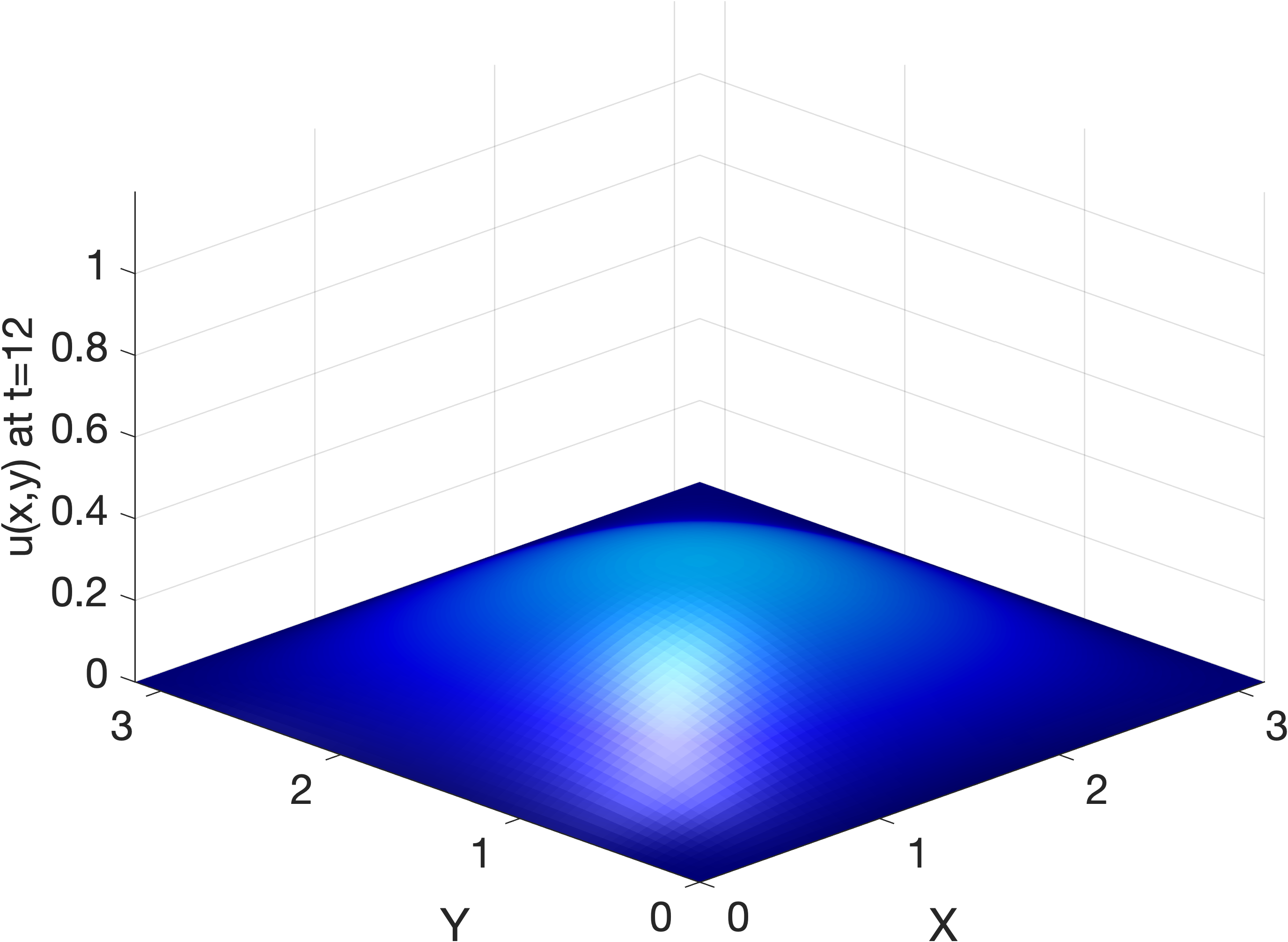}
			\end{minipage}
		}
		\caption{\label{eg3_numerical_solution} Numerical Solutions of 2D wave equation at Different Time Points on t=0, t=6 and t=12.}
	\end{figure}

\begin{figure}[p]
		\centering
		\subfigure{
			\centering
			\includegraphics[width=0.85\textwidth]{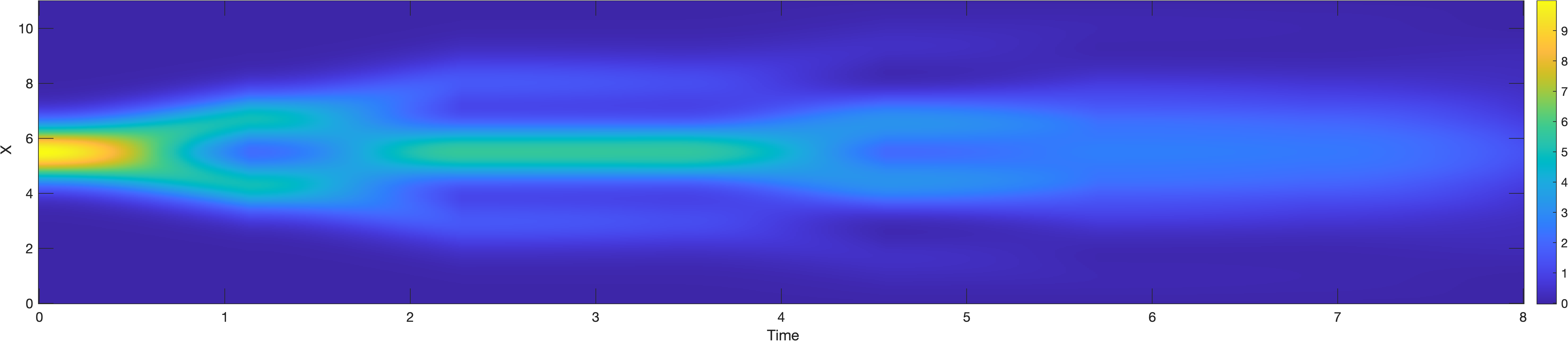}
		}
		
		\subfigure{
			\centering
			\includegraphics[width=0.85\textwidth]{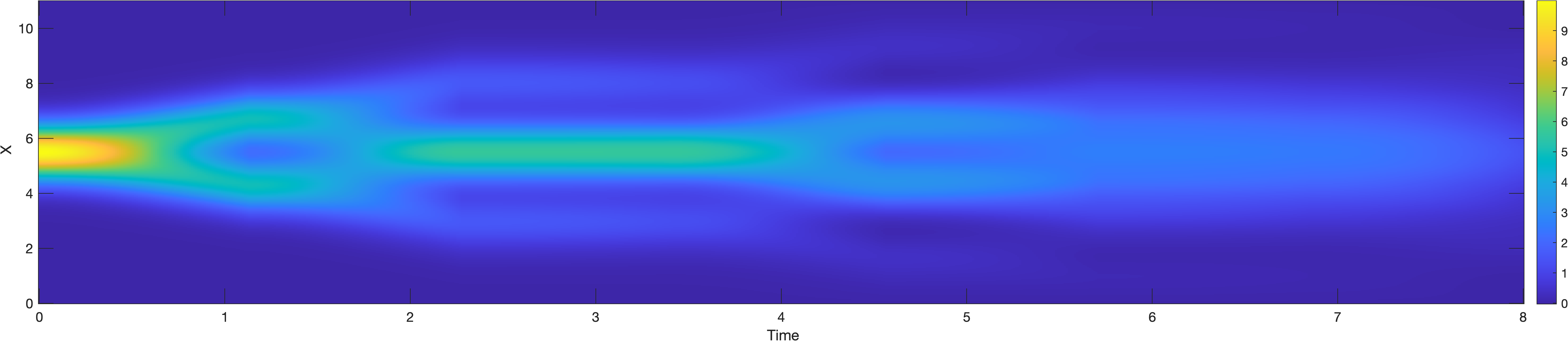}
		}
		
		\subfigure{
			\centering
			\includegraphics[width=0.855\textwidth]{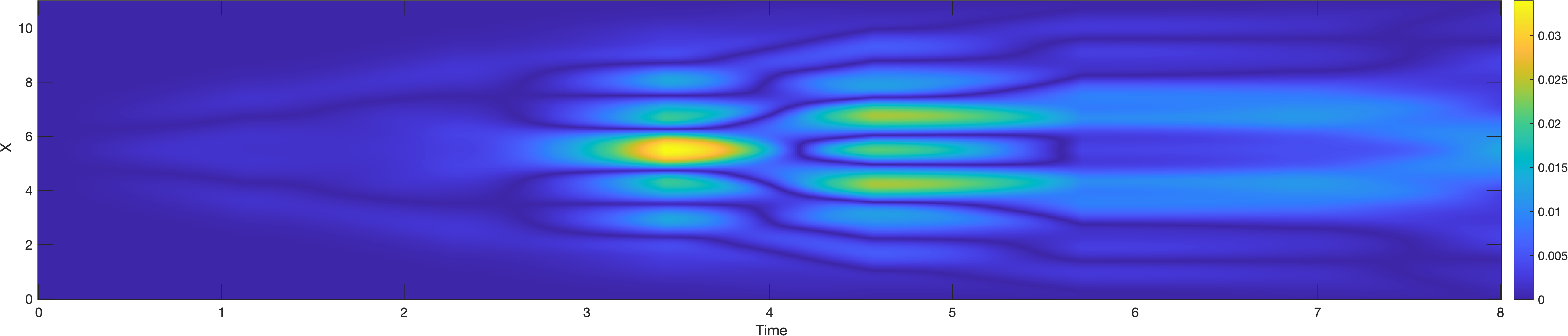}
		}
		\caption{\label{SG_heat_map3} Numerical Solution, Inferred Solution and Error Heat Map of SG Equation with Markovian Switching}
		
		\subfigure{
			\centering
			\includegraphics[width=0.85\textwidth]{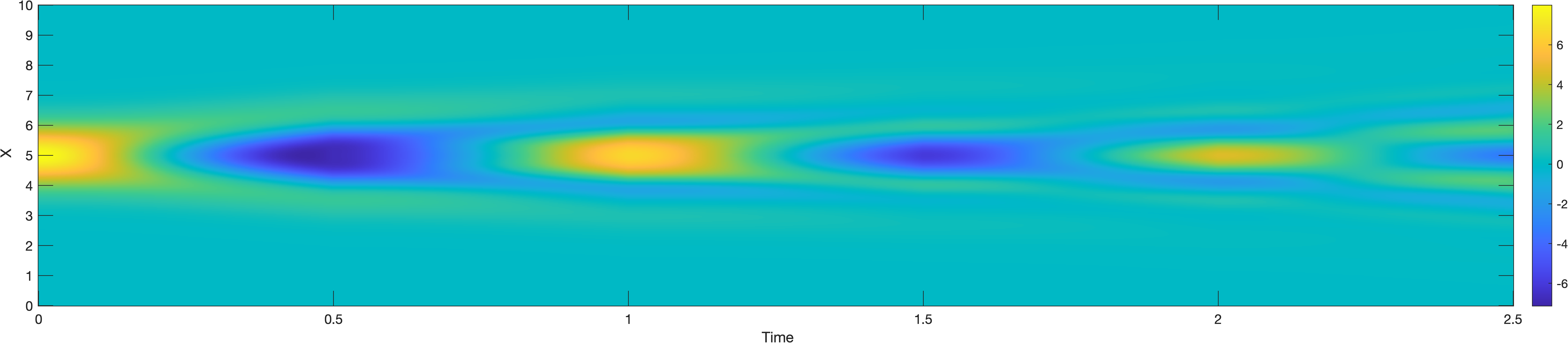}
		}
		
		\subfigure{
			\centering
			\includegraphics[width=0.85\textwidth]{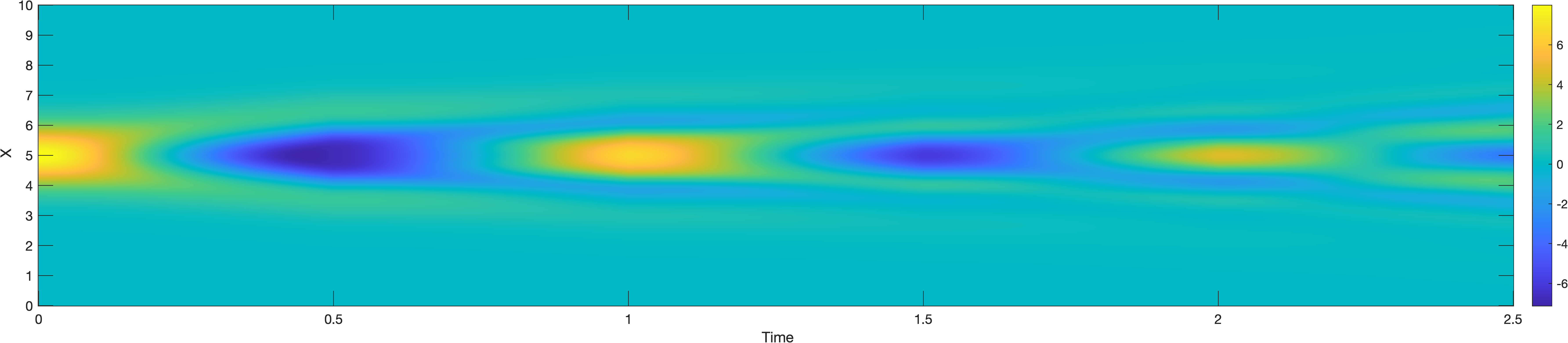}
		}
		
		\subfigure{
			\centering
			\includegraphics[width=0.855\textwidth]{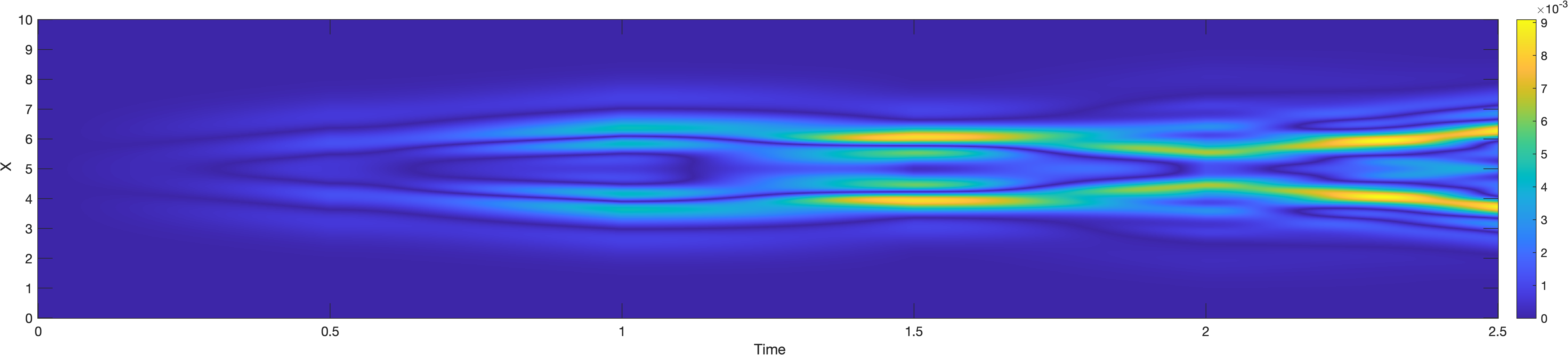}
		}
		\caption{\label{KG_heat_map3} Numerical Solution, Inferred Solution and Error Heat Map of KG Equation with Markovian Switching}
	\end{figure}

	The parameter inference results of the KG equation with Markovian switching are shown in Table \ref{tab:KG}. The visualization of these results is shown in Fig.\ref{KG_solution}. Moreover, Fig.~\ref{KG_heat_map3} displays the error heat map, illustrating the discrepancies between the numerical and inferred solutions, with the magnitude around \(10^{-3}\), indicating a high degree of accuracy in the parameter estimation.
	
	\subsection{Case 3: A two-dimensional nonlinear wave equation with Markovian switching}
	The preceding two cases are both one-dimensional. Now we consider a two-dimensional nonhomogeneous wave equation with Markovian switching. Moreover, we compare the performance of the mixture model and the single model which refers to the process of applying generated data to traditional parameter estimation methods, such as regularization techniques or the least squares method. Here, the comparison is made with the standard SBL algorithm. The specific form of the equation is given as follows
	\begin{equation}{\label{eq:2D_wave}}
		\begin{cases}
			u_{tt}(\mathbf{x}, t) - M_t \Delta u = f(\mathbf{x}, t), & \text{in } U_T \\
			u(\mathbf{x}, 0) = g(\mathbf{x}), \; u_t(\mathbf{x}, 0) = h(\mathbf{x}), & \text{on } U \times \{t=0\} \\
			u(\mathbf{x}, t) = 0, & \text{on } \partial U \times [0, T],
		\end{cases}
	\end{equation}
	where \(U = [0, \pi] \times [0, \pi]\), \(T = 20\). The initial and boundary conditions are determined by the exact solutions
	\begin{equation}
		u(x, y, t) = e^{-t} \sin x \sin y,
	\end{equation}
	and \(f\) is defined as
	\begin{equation}
		f(x, y, t) = (1 + 2M_{t_i}) e^{-t} \sin x \sin y, \quad t \in [t_i, t_{i+1}].
	\end{equation}
	
	We define a function $\mathcal{J}$ that maps the time interval to a finite set of system parameter changes
	\begin{equation}
		\mathcal{J}: [0,T] \rightarrow \{\theta_1, ..., \theta_n\}
	\end{equation}
	$\mathcal{J}$ represents the jump points of the parameters controlled by the Markov chain $M_t$ over a given period. In this case, the jump behavior of PDEs' parameter governed by the Markov chain is given by
	\begin{equation} 
		\mathcal{J}(t_i) = \{1, 2, 3, 2, 1, 3, 2, 1, 2, 3\},\quad i=0,1,2,...,10, \ t_0=0,
	\end{equation}
	where $t_i$ denotes the occurrence times of the jumps.
	
	\begin{table}[t]
		\centering
		\begin{tabular}{c c c}
			\toprule
			& Time Interval & 2D Wave Equation(\textcolor{red}{Absolute error)} \\
			\midrule
			Simulation Set & $[0,t_1)\cup[t_4,t_5)\cup [t_7,t_8)$ & \( u_{tt} = 1u_{xx} + f(x,y,t) \) \\
			& $[t_1,t_2)\cup[t_3,t_4)\cup[t_6,t_7)\cup[t_8,t_9)$ & \( u_{tt} = 2u_{xx} + f(x,y,t) \) \\
			& $[t_2,t_3)\cup[t_5,t_6)\cup[t_9,t_{10}]$ & \( u_{tt} = 3u_{xx} + f(x,y,t) \) \\
			\midrule
			\textbf{Mixture Model} & $[0,t_1)\cup[t_4,t_5)\cup [t_7,t_8)$ & \( u_{tt} = 0.9832 \textcolor{red}{(1.68\%)} u_{xx} + f(x,y,t) \) \\
			& $[t_1,t_2)\cup[t_3,t_4)\cup[t_6,t_7)\cup[t_8,t_9)$ & \( u_{tt} = 1.9801 \textcolor{red}{(0.95\%)} u_{xx} + f(x,y,t) \) \\
			& $[t_2,t_3)\cup[t_5,t_6)\cup[t_9,t_{10}]$ & \( u_{tt} = 3.0426 \textcolor{red}{(1.42\%)} u_{xx} + f(x,y,t) \) \\
			\midrule
			Single Model & $[0,t_1)\cup[t_4,t_5)\cup [t_7,t_8)$ & \( u_{tt} = 2.2293 \textcolor{red}{(122.93\%)} u_{xx} + f(x,y,t) \) \\
			& $[t_1,t_2)\cup[t_3,t_4)\cup[t_6,t_7)\cup[t_8,t_9)$ & \( u_{tt} = 2.2293 \textcolor{red}{(11.465\%)} u_{xx} + f(x,y,t) \) \\
			& $[t_2,t_3)\cup[t_5,t_6)\cup[t_9,t_{10}]$ & \( u_{tt} = 2.2293 \textcolor{red}{(25.690\%)} u_{xx} + f(x,y,t) \) \\
			\bottomrule
		\end{tabular}
		\caption{\label{tab:wave_eq} Parameter inference results of the DSBL algorithm applied to the wave equation with different models.}
	\end{table}

	To generate the corresponding spatiotemporal data, we employ a numerical method with non-uniform grids. We use a basic uniform grid with a time step of 0.01 and a spatial step of 1/64, resulting in a dataset of size \(65 \times 65 \times 2001\). Since the equation is two-dimensional, we can only visualize the numerical solution at specific times, as illustrated in Fig.~\ref{eg3_numerical_solution}.

	\begin{figure}[p]
		\centering
		\subfigure{
			\begin{minipage}{0.195\textwidth}
				\centering
				\includegraphics[width=1\textwidth]{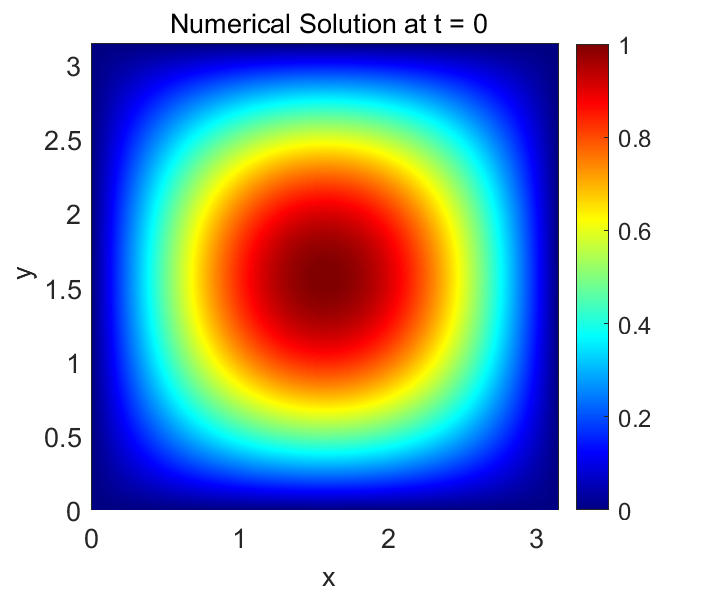}
			\end{minipage}
			\begin{minipage}{0.195\textwidth}
				\centering
				\includegraphics[width=1\textwidth]{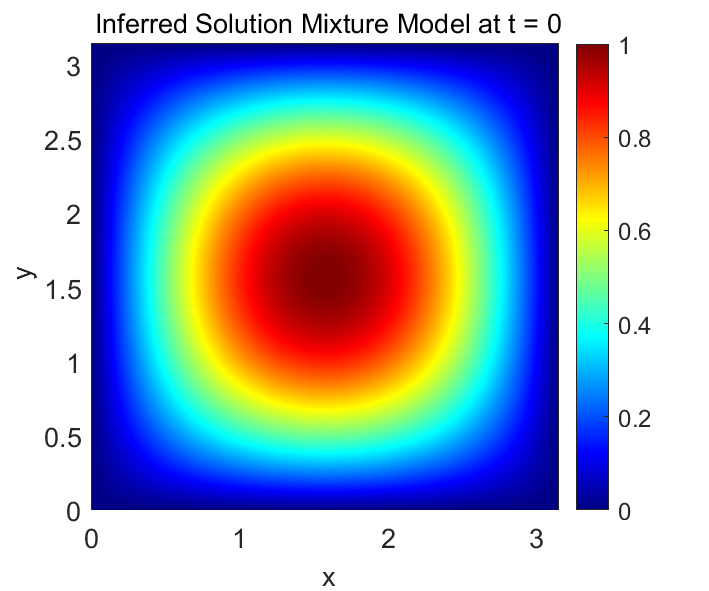}
			\end{minipage}
			\begin{minipage}{0.195\textwidth}
				\centering
				\includegraphics[width=1\textwidth]{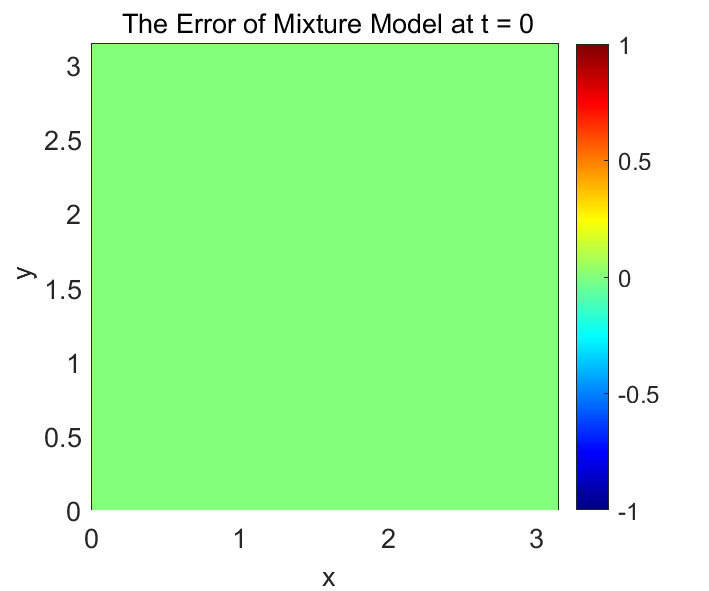}
			\end{minipage}
			\begin{minipage}{0.195\textwidth}
				\centering
				\includegraphics[width=1\textwidth]{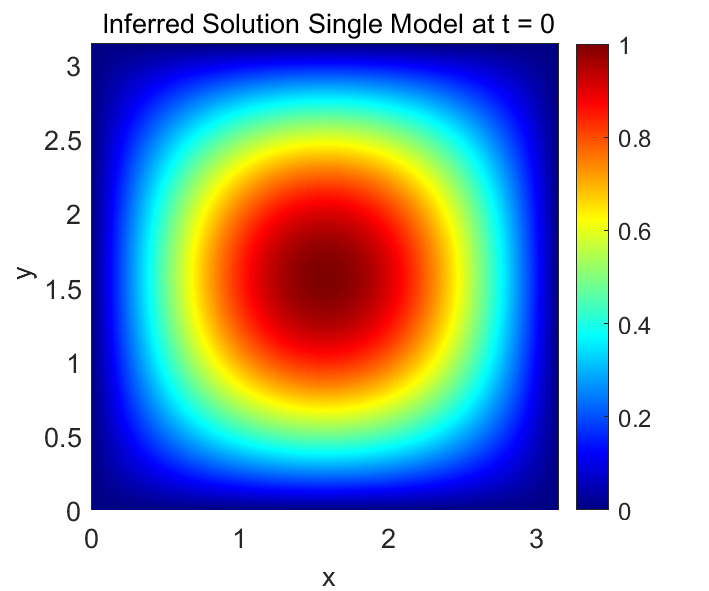}
			\end{minipage}
			\begin{minipage}{0.195\textwidth}
				\centering
				\includegraphics[width=1\textwidth]{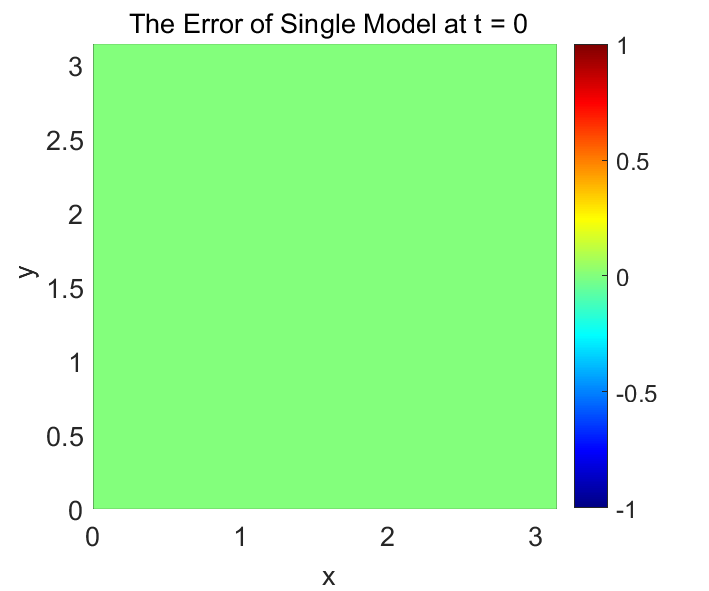}
			\end{minipage}
		}
		\subfigure{
			\begin{minipage}{0.195\textwidth}
				\centering
				\includegraphics[width=1\textwidth]{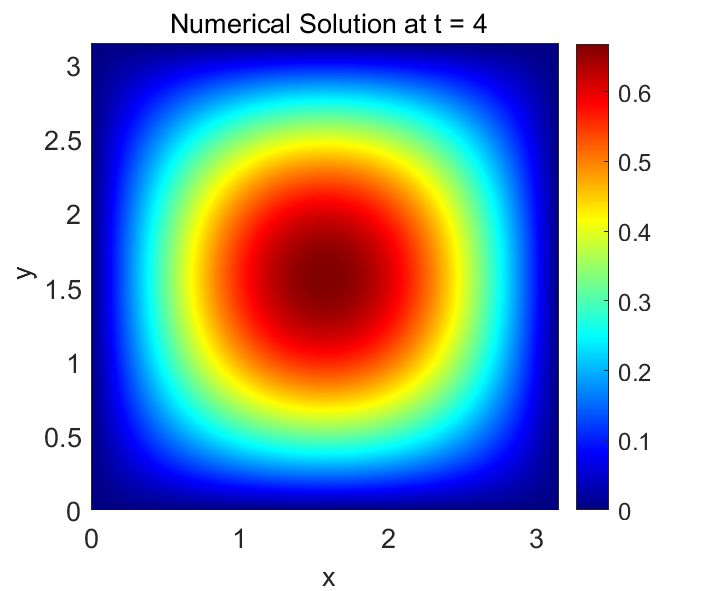}
			\end{minipage}
			\begin{minipage}{0.195\textwidth}
				\centering
				\includegraphics[width=1\textwidth]{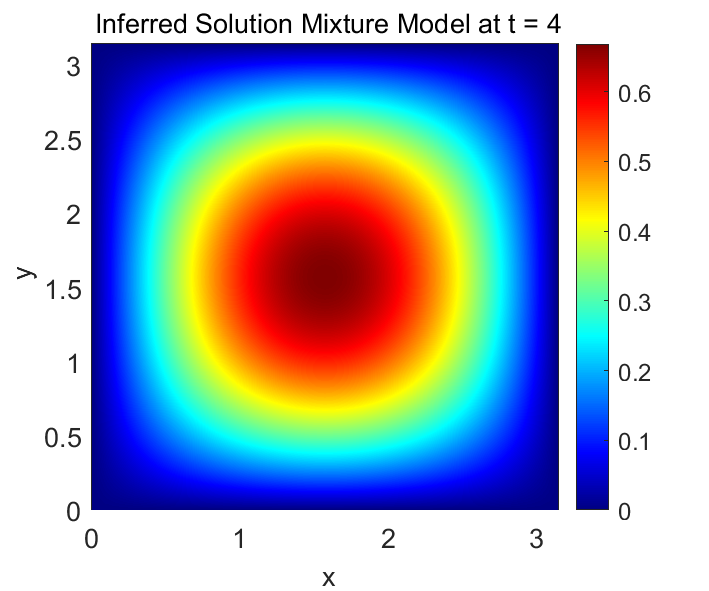}
			\end{minipage}
			\begin{minipage}{0.195\textwidth}
				\centering
				\includegraphics[width=1\textwidth]{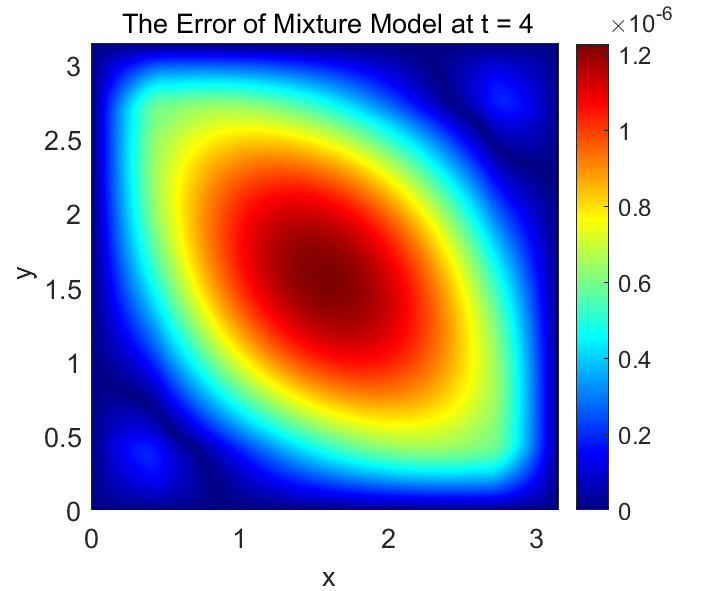}
			\end{minipage}
			\begin{minipage}{0.195\textwidth}
				\centering
				\includegraphics[width=1\textwidth]{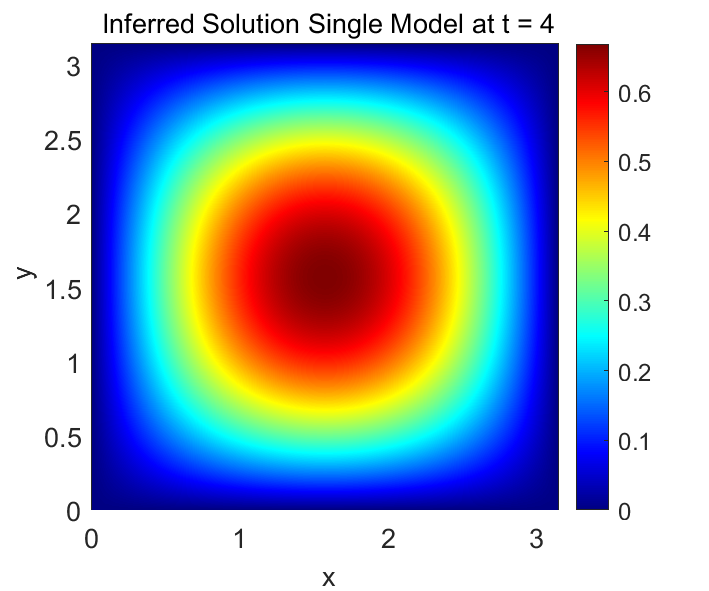}
			\end{minipage}
			\begin{minipage}{0.195\textwidth}
				\centering
				\includegraphics[width=1\textwidth]{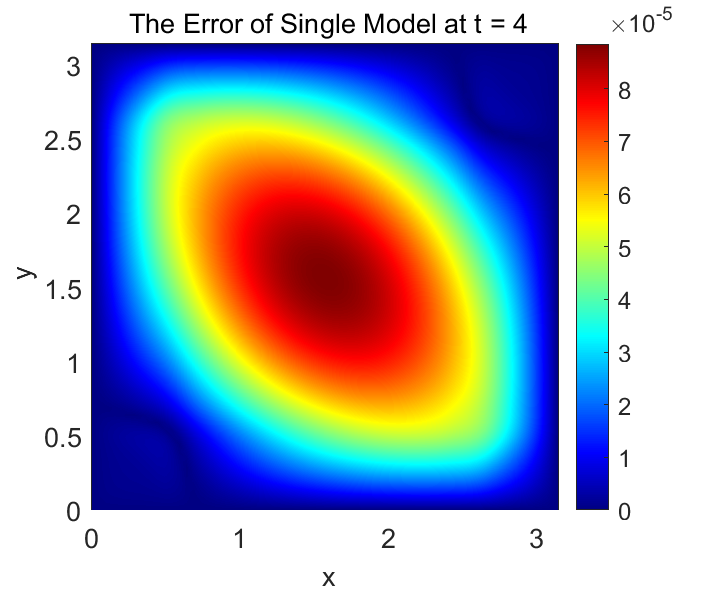}
			\end{minipage}
		}
		\subfigure{
			\begin{minipage}{0.195\textwidth}
				\centering
				\includegraphics[width=1\textwidth]{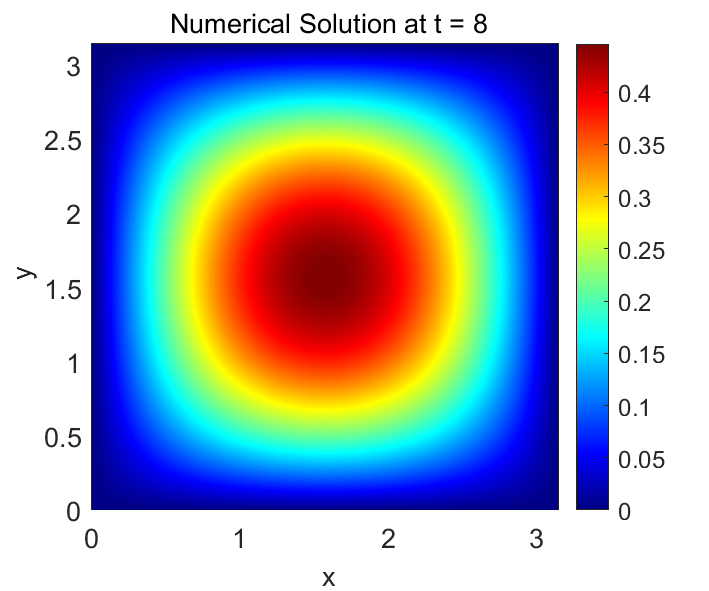}
			\end{minipage}
			\begin{minipage}{0.195\textwidth}
				\centering
				\includegraphics[width=1\textwidth]{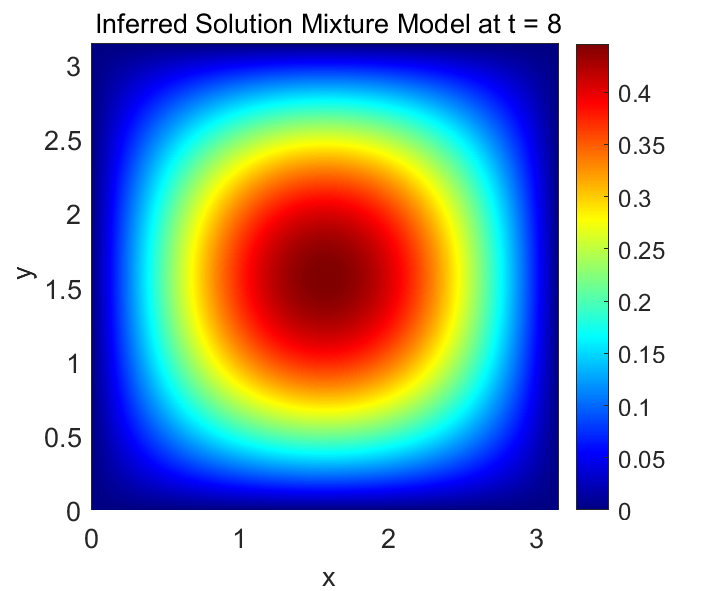}
			\end{minipage}
			\begin{minipage}{0.195\textwidth}
				\centering
				\includegraphics[width=1\textwidth]{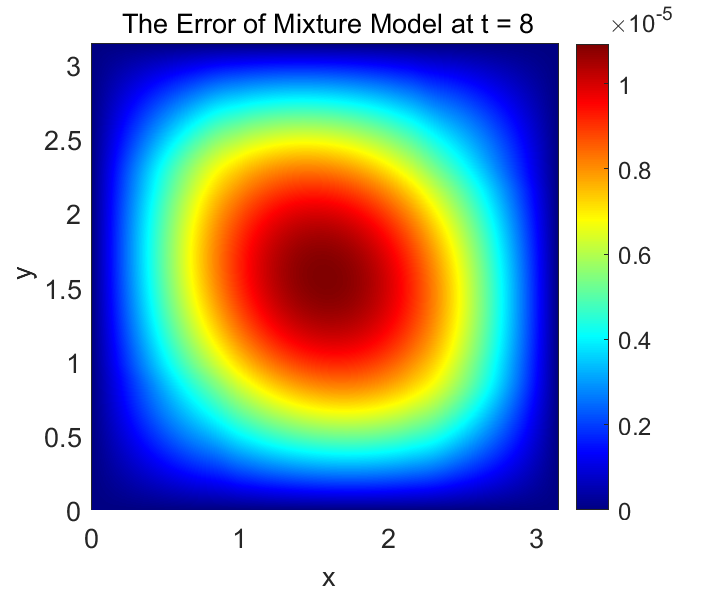}
			\end{minipage}
			\begin{minipage}{0.195\textwidth}
				\centering
				\includegraphics[width=1\textwidth]{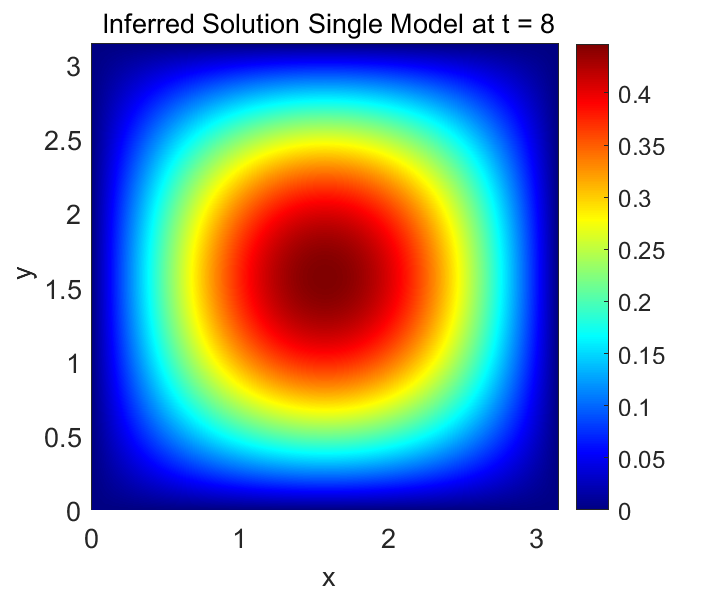}
			\end{minipage}
			\begin{minipage}{0.195\textwidth}
				\centering
				\includegraphics[width=1\textwidth]{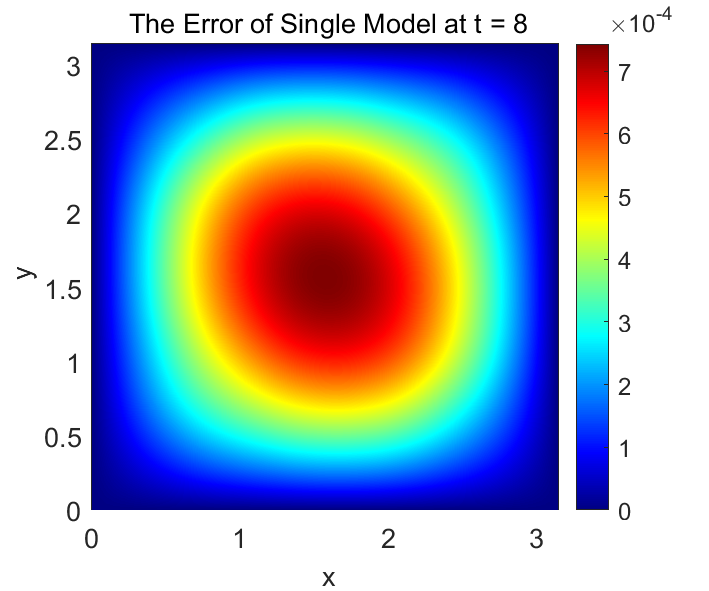}
			\end{minipage}
		}
		\subfigure{
			\begin{minipage}{0.195\textwidth}
				\centering
				\includegraphics[width=1\textwidth]{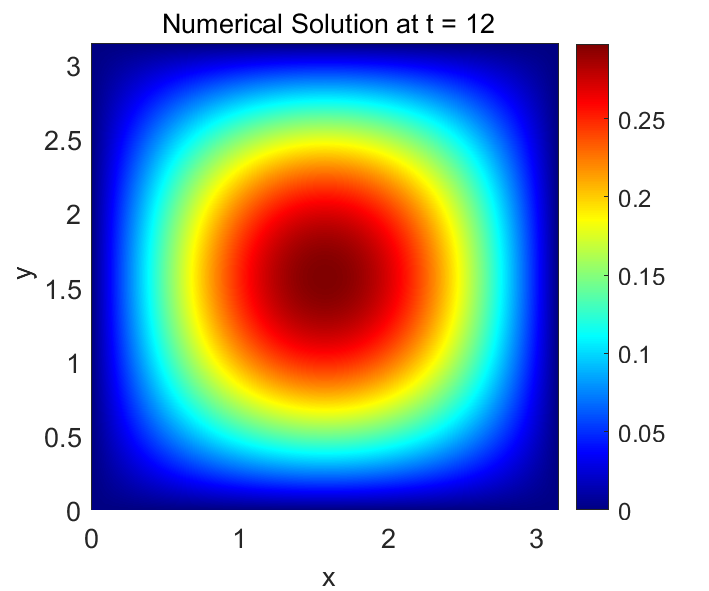}
			\end{minipage}
			\begin{minipage}{0.195\textwidth}
				\centering
				\includegraphics[width=1\textwidth]{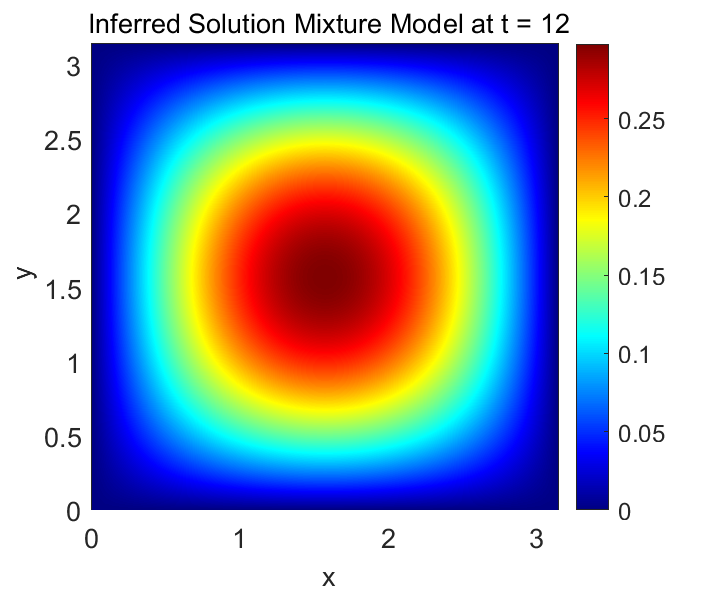}
			\end{minipage}
			\begin{minipage}{0.195\textwidth}
				\centering
				\includegraphics[width=1\textwidth]{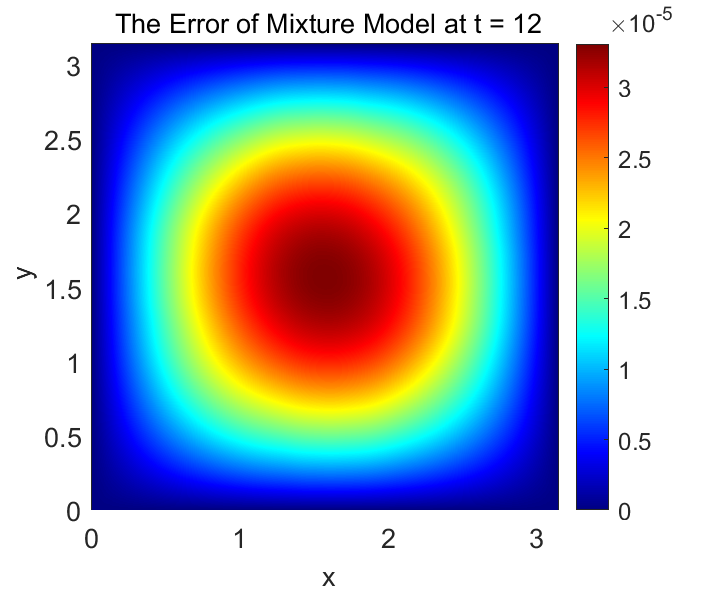}
			\end{minipage}
			\begin{minipage}{0.195\textwidth}
				\centering
				\includegraphics[width=1\textwidth]{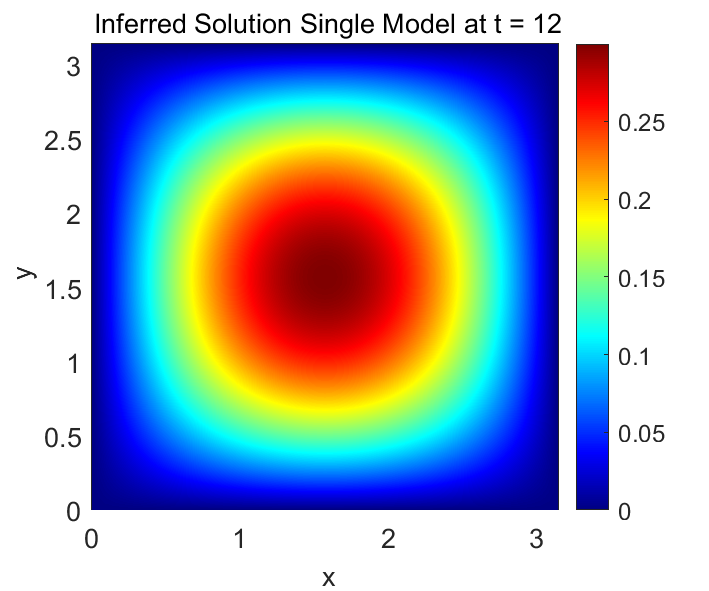}
			\end{minipage}
			\begin{minipage}{0.195\textwidth}
				\centering
				\includegraphics[width=1\textwidth]{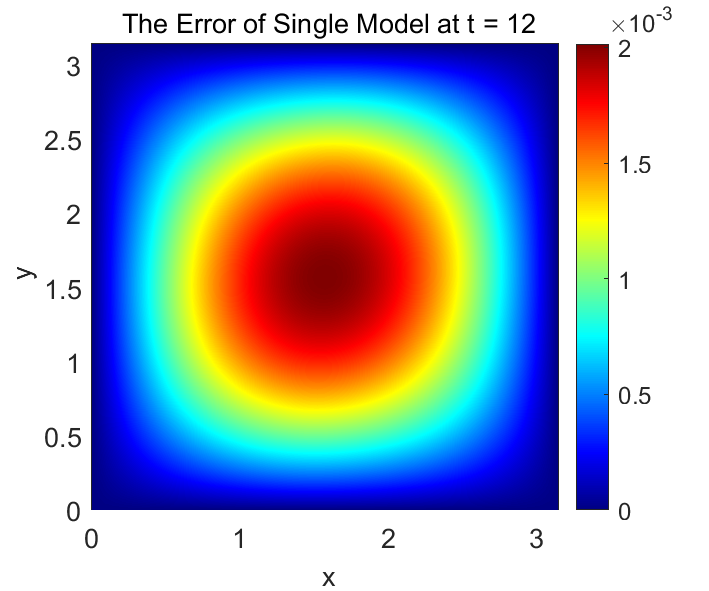}
			\end{minipage}
		}
		\subfigure{
			\begin{minipage}{0.195\textwidth}
				\centering
				\includegraphics[width=1\textwidth]{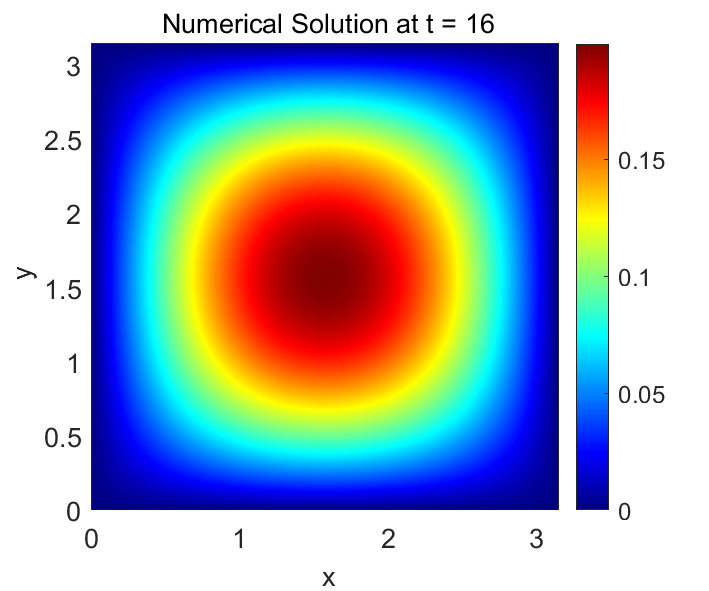}
			\end{minipage}
			\begin{minipage}{0.195\textwidth}
				\centering
				\includegraphics[width=1\textwidth]{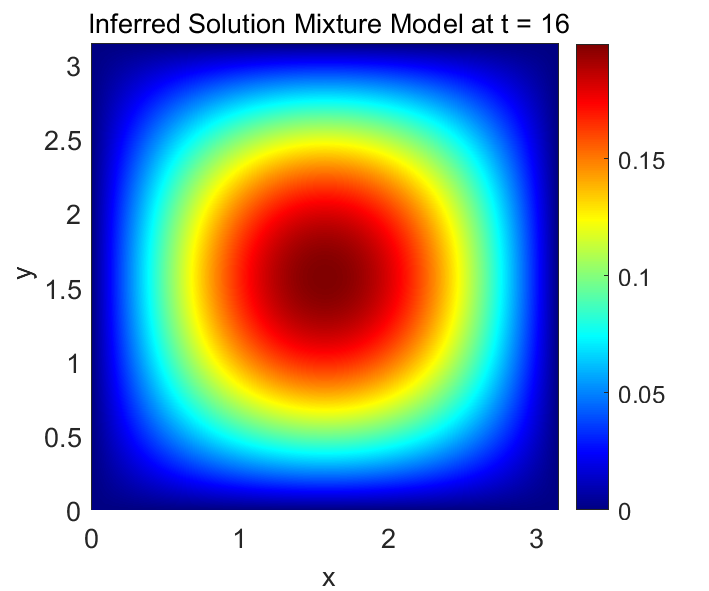}
			\end{minipage}
			\begin{minipage}{0.195\textwidth}
				\centering
				\includegraphics[width=1\textwidth]{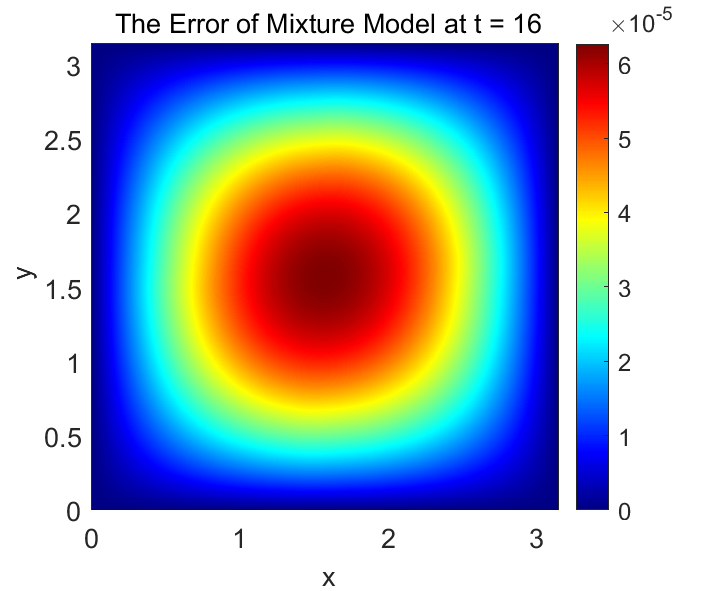}
			\end{minipage}
			\begin{minipage}{0.195\textwidth}
				\centering
				\includegraphics[width=1\textwidth]{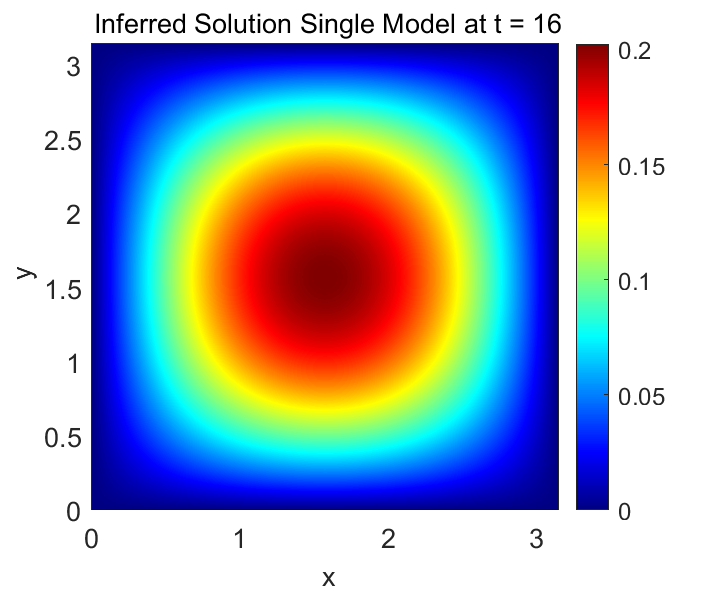}
			\end{minipage}
			\begin{minipage}{0.195\textwidth}
				\centering
				\includegraphics[width=1\textwidth]{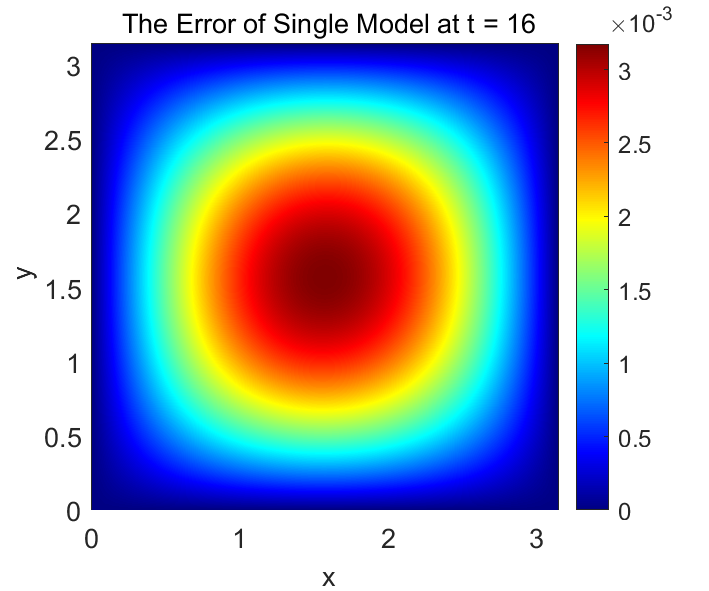}
			\end{minipage}
		}
		\subfigure{
			\begin{minipage}{0.195\textwidth}
				\centering
				\includegraphics[width=1\textwidth]{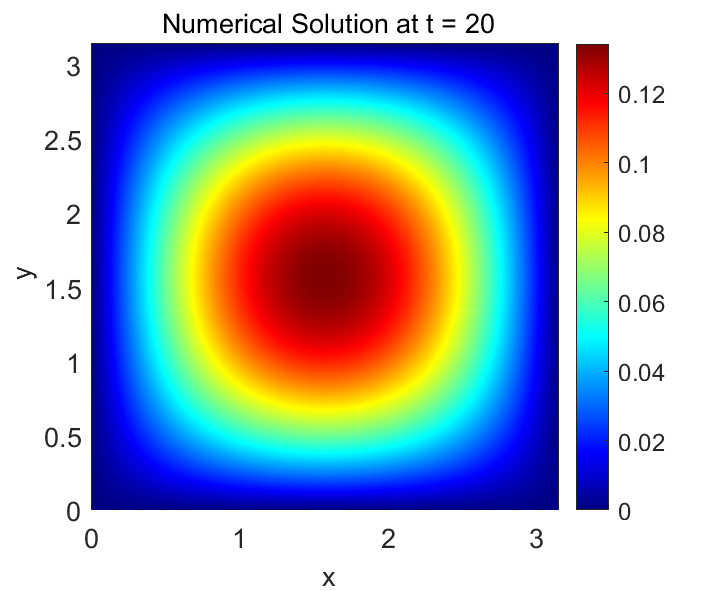}
			\end{minipage}
			\begin{minipage}{0.195\textwidth}
				\centering
				\includegraphics[width=1\textwidth]{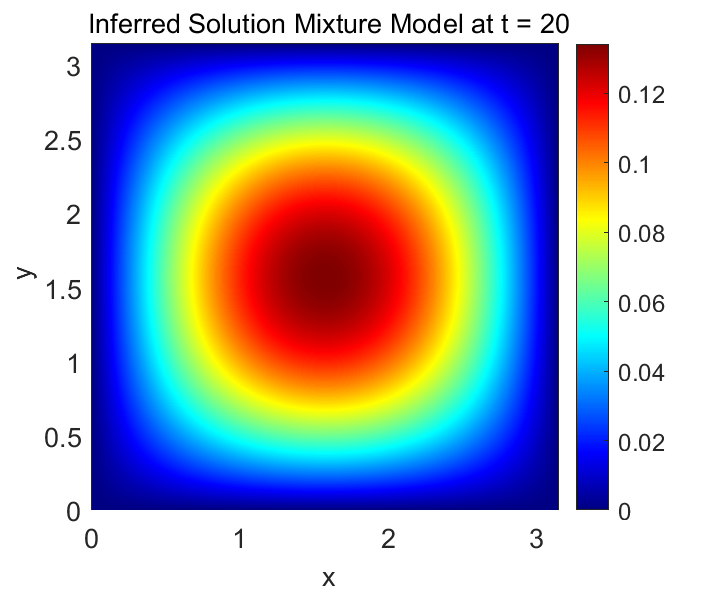}
			\end{minipage}
			\begin{minipage}{0.195\textwidth}
				\centering
				\includegraphics[width=1\textwidth]{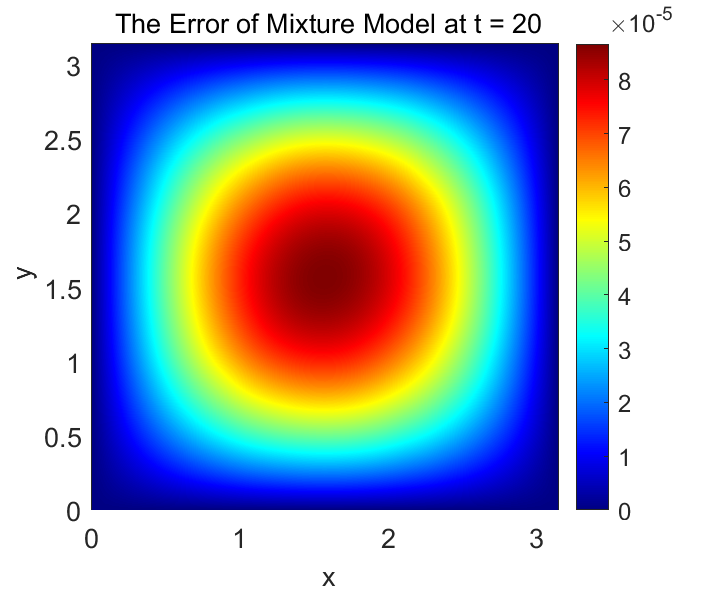}
			\end{minipage}
			\begin{minipage}{0.195\textwidth}
				\centering
				\includegraphics[width=1\textwidth]{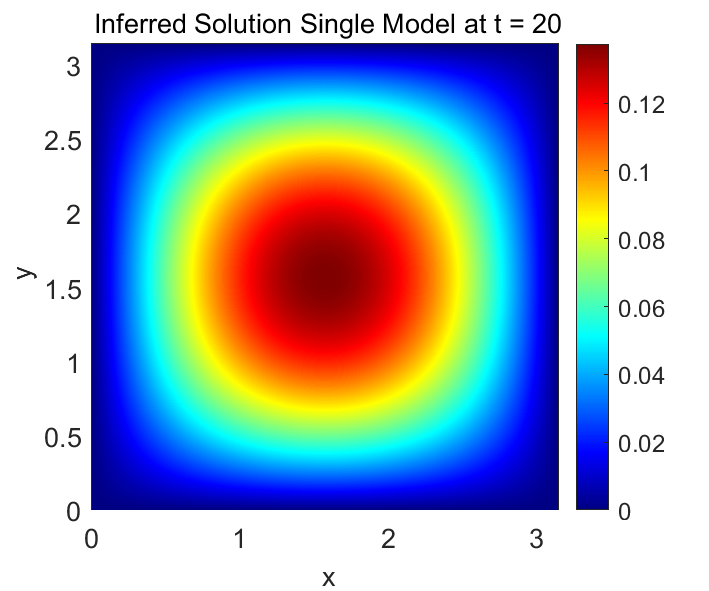}
			\end{minipage}
			\begin{minipage}{0.195\textwidth}
				\centering
				\includegraphics[width=1\textwidth]{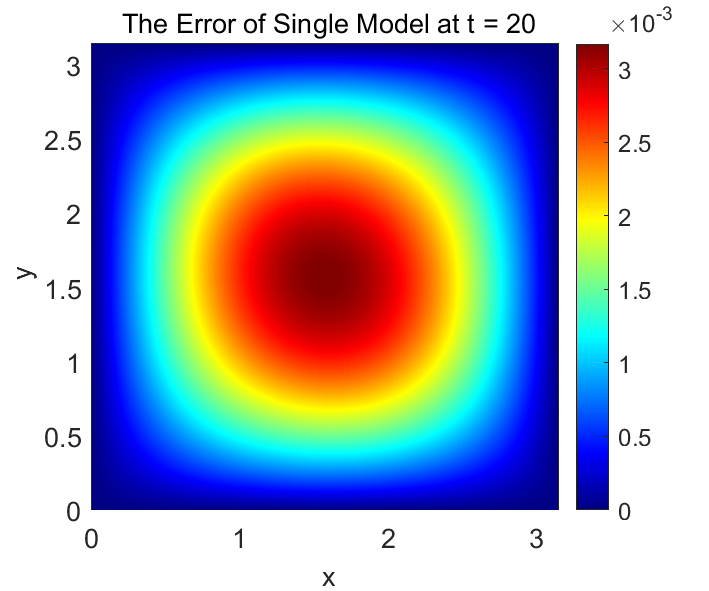}
			\end{minipage}
		}
		\caption{\label{2D_wave} Numerical vs. Inferred Solution and Corresponding Error Heat Map for Mixture Model and Single Model}
	\end{figure}

	The shape of the wave changes over time. After applying the DSBL and SBL algorithm to the spatiotemporal dataset respectively, the parameter inference results are presented in Table.~\ref{tab:wave_eq}. The mixture model using DSBL demonstrates high accuracy and good adaptability to the switching of system coefficients.
	
	Additionally, several solutions over time are visualized in Fig.~\ref{2D_wave}. The first column represents the evolution of the numerical solution \(u(x,t)\) of the mixture model in the simulation experiment over time. The second and fourth columns show the inferred solutions of the model parameters using the DSBL algorithm and the standard SBL algorithm, respectively, as they evolve over time. The third and fifth columns display the absolute errors between the inferred solutions and the numerical solutions obtained using these two methods. It can be observed that, at the beginning of the experiment, both methods handle the wave equation data well. However, the absolute error of the SBL method becomes increasingly larger compared to our method, leading to a decline in estimation accuracy. This may be due to the accumulation of estimation biases over a longer period.
	
	\section{Conclusion and future work}
	This paper focuses on parameter inference for a newly proposed wave equation with Markovian switching and introduces a Discrete Sparse Bayesian Learning
	algorithm specifically designed for such systems. To better simulate real-world phenomena involving abrupt changes, it is essential to accurately capture the dynamic variations of the system across spatial dimensions. In this approach, the coefficients of the wave system are modeled as a Markov process. We establish theorems for the convergence and error bounds of the DSBL algorithm, providing a solid theoretical foundation for its application. Three explicit numerical experiments are conducted to validate the effectiveness of the proposed algorithm. The results demonstrate that the DSBL algorithm offers superior accuracy in fitting data with noise generated by the systems, outperforming traditional methods in parameter estimation. The robustness of the DSBL algorithm in handling noisy data underscores its advantage in accurately inferring parameters for complex systems with sudden changes.
	
	Looking ahead, integrating the DSBL method into deep learning frameworks or combining it with other machine learning techniques could enhance the speed and accuracy of inferred solutions. Further research could focus on optimizing the selection of prior knowledge in Bayesian methods and improving iterative formats for parameter estimation within difference-of-convex programming, which could lead to greater accuracy and robustness in solving complex, ill-posed partial differential equations. Additionally, exploring real-time parameter identification algorithms for hybrid system states and studying error distributions beyond the current scope could open new avenues for advancement in this field.
	
	\section*{Acknowledgements}
	This work was supported by the NSFC grant 12371198 and the fundamental research funds for the central universities under grand YCJJ20242224.
	
	\section{Appendix}\label{Appendix}
	\subsection{A proof of Theorem~\ref{bounded sequence}}\label{Appendix0}
	$Proof:$ Before presenting the proof of the theorem, several assumptions are provided. The columns of \( D \) are nonzero (i.e., \( D_i \neq 0 \) for \( i=1, \ldots, m \)), and if \( \theta_i \) and \( \gamma_i \) converge to zero simultaneously, then only the path in the space \((\theta, \gamma)\) of the form as
	\begin{equation}
		\theta_i = \mathcal{O}(\gamma_i), \quad \text { as }   \gamma_i \rightarrow 0^{+},
	\end{equation}
	which will be considered. This ensures the limit of loss function \(\mathcal{L}\) exists along the path \((\theta, \gamma)\). Define the set \(\Omega = \mathbb{R}^m \times [0, \infty)^m\) and note that \(\mathcal{L}\) is only defined on \(\text{int}(\Omega)\). Therefore, we need to extend it to \(\bar{\Omega} = \text{int}(\Omega) \cup \partial \Omega\).
	
	For any point \((\theta, \gamma) \in \partial \Omega\), if \(\theta_i\) and \(\gamma_i\) are both zero or if \(\gamma_i \neq 0\) for any \(i \in \{1, 2, \cdots, m\}\), the limit of \(\mathcal{L}\) at the point \((\theta, \gamma)\) exists and is unique along the path satisfying the condition above. This limit is referred to as the value of the function \(\mathcal{L}\) at the point \((\theta, \gamma)\). If there exists an \(i\) such that \(\theta_i \neq 0\) and \(\gamma_i = 0\), then the limit of the function \(\mathcal{L}\) at the point \((\theta, \gamma)\) along the path satisfying the condition above is positive infinity. In this case, we define \(\mathcal{L}(\theta, \gamma) = M\), where \(M\) is a sufficiently large number.
	
	The boundary set can be divided into two disjoint sets, i.e., \(\partial \Omega = \Omega_1 \cup \Omega_2\), and we define
	\begin{equation}
		\Omega_1 := \left\{(\theta, \gamma) \mid \text{there exists at least one } i \text{ such that } \theta_i \neq 0 \text{ and } \gamma_i = 0\right\}, \quad \Omega_2 := \partial \Omega \backslash \Omega_1.
	\end{equation}
	
	The concrete form of the loss function is  
	\begin{equation}\label{loss_L}
		\mathcal{L}(\theta, \gamma)  =\frac{1}{\sigma^{2}}\|y-D \theta\|^{2}+\theta^{T} \Gamma^{-1} \theta+\log \left|\sigma^{2} I_{n}+D \Gamma D^{T}\right| ,
	\end{equation}
	and bounded non-increasing sequence $\{\mathcal{L}(\theta^{(k)},\gamma^{(k)})\}_{k=0}^{\infty}$ is obtained by
	\begin{equation}\label{Appendix_loss}
		(\theta^{(k+1)}, \gamma^{(k+1)}) \in \argmin_{(\theta, \gamma) \in \Omega} \hat{\mathcal{L}}(\theta, \gamma; \theta^{(k)}, \gamma^{(k)}).
	\end{equation}
	The conditional loss function is
	\begin{equation}
		\widehat{\mathcal{L}}\left(\theta, \gamma ; \theta^{(k)}, \gamma^{(k)}\right)  =f(\theta, \gamma)+g\left(\theta^{(k)}, \gamma^{(k)}\right)+
		\nabla_{\gamma}g\left(\theta^{(k)}, \gamma^{(k)}\right)^{T}(\gamma-\gamma^{(k)}),
	\end{equation}
	where
	\begin{equation}
		c^{(k)} =\nabla_{\gamma} \log \left|\sigma^{2} I_{n}+D \Gamma^{(k)} D^{T}\right|, \quad \Gamma^{(k)} =\operatorname{diag}\left(\gamma_{1}^{(k)}, \gamma_{2}^{(k)}, \cdots, \gamma_{m}^{(k)}\right).
	\end{equation}
	The above relation is equal to
	\begin{equation}
		\left(\theta^{(k+1)}, \gamma^{(k+1)}\right) \in \arg \min _{(\theta, \gamma) \in \Omega}\left\{\frac{1}{\sigma^{2}}\|y-D \theta\|^{2}+\sum_{i}\left(\frac{\theta_{i}^{2}}{\gamma_{i}}+c_{i}^{(k)} \gamma_{i}\right)\right\},
	\end{equation}
	
	Note that the supplementary definition of $\mathcal{L}$ can be extended to the function $\widehat{\mathcal{L}}$. From this, it follows that
	\begin{equation}\label{19}
		\theta^{(k+1)} \in \arg \min _{\theta}\left\{\|y-D \theta\|^{2}+2 \sigma^{2} \sum_{i} \sqrt{c_{i}^{(k)}}\left|\theta_{i}\right|\right\}, \quad
		\gamma_{i}^{(k+1)}=\frac{\left|\theta_{i}^{(k+1)}\right|}{\sqrt{c_{i}^{(k)}}}, \quad i=1,2, \cdots, m.
	\end{equation}
	
	Given an initial point $\left(\theta^{(0)}, \gamma^{(0)}\right) \in \operatorname{int}(\Omega)$, an iterative sequence $\left\{\left(\theta^{(k)}, \gamma^{(k)}\right)\right\}_{k=0}^{\infty}$ is obtained. Additionally, $c^{(k)}$ can be computed analytically as
	\begin{equation}\label{20}
		c^{(k)} = \operatorname{diag}\left[D^{T}\left(\sigma^{2} I_{n}+D \Gamma^{(k)} D^{T}\right)^{-1} D\right].
	\end{equation}
	
	By leveraging the properties of concave functions, it can be deduced that
	\begin{equation}
		\mathcal{L}(\theta, \gamma) \leq \widehat{\mathcal{L}}\left(\theta, \gamma ; \theta^{(k)}, \gamma^{(k)}\right).
	\end{equation}
	
	Combining this with Eq.\ref{CCP}, we have
	\begin{equation}
		\begin{aligned}
			&\mathcal{L}\left(\theta^{(k+1)}, \gamma^{(k+1)}\right)  \leq \widehat{\mathcal{L}}\left(\theta^{(k+1)}, \gamma^{(k+1)} ; \theta^{(k)}, \gamma^{(k)}\right) \\
			\leq & \widehat{\mathcal{L}}\left(\theta^{(k)}, \gamma^{(k)} ; \theta^{(k)}, \gamma^{(k)}\right)
			= \mathcal{L}\left(\theta^{(k)}, \gamma^{(k)}\right).
		\end{aligned}
	\end{equation}
	
	Therefore, it follows that
	\begin{equation}
		\mathcal{L}\left(\theta^{(k)}, \gamma^{(k)}\right) \geq \log \left|\sigma^{2} I_{n}\right|=2 n \log \sigma>-\infty.
	\end{equation}
	
	In conclusion, the sequence $\left\{\mathcal{L}\left(\theta^{(k)}, \gamma^{(k)}\right)\right\}_{k=0}^{\infty}$ is bounded and monotonically non-increasing. This completes the proof of the theorem.
	
	\subsection{A proof of Theorem~\ref{convergence}}\label{Appendix1}
	$Proof:$ When $M_t$ is continuous, let $\mathcal{I}: [0,T]\to \mathbb{R}$. It is called \textit{c\`adl\`ag} if for all $t\in [0,T]$, $\mathcal{I}(t)$ has a left limit and is right-continuous at $t$. Since the irreducible and aperiodic $M_t$ has finite states and all states are stationary, almost all sample paths of $M_t$ are \textit{c\`adl\`ag} functions, as the chain experiences a finite number of jumps in any finite time interval.
	
	For every \textit{c\`adl\`ag} function $\mathcal{I}(t)$  defined on $[0, T]$ and for every $ k > 0$, there exists a finite partition $\mathcal{P} = 0 < t_1 < \cdots < t_n = T $ such that
	\begin{equation}\label{proof_1}
		\sup\{|\mathcal{I}(u)-\mathcal{I}(v)|: u,v\in [t_j,t_{j+1}),j=0,1,...,n-1\}<k
	\end{equation}
	If such a partition does not exist over the global time interval $[0,T]$, we fix $k > 0$. Nonetheless, there still exist some left-closed right-open intervals $[0,s)$ with $s \in [0,T]$ that satisfy the finite partition condition Eq.\ref{proof_1}. Let $t^*$ be the supremum of the set of all such possible points $s$. Since $\mathcal{I}(t)$ is right-continuous at 0, we have $t^* > 0$. Furthermore, since $\mathcal{I}(t)$ has a left limit at $t^*$, the interval $[0, t^*)$ can be partitioned as required. Suppose $t^* < T$. Given the right-continuity of $\mathcal{I}(t)$ at $t^*$, there exists $t^{**} > t^*$ such that for $u, v \in [t^*, t^{**})$, $\sup |I(u) - I(v)| < k$. This contradicts the definition of \( t^* \). Therefore, we conclude that $t^* = T$.
	
	Next, we define the jump of the sample path \(\mathcal{I}(t)\) of \(M_t\) at \(t\) as
	\begin{equation}
		\Delta \mathcal{I}(t) = \mathcal{I}(t) - \mathcal{I}(t-).
	\end{equation}
	
	For a finite partition $\mathcal{P}$ of a sample path of the continuous-time Markov chain $M_t$, note that the discontinuities in the sample of $M_t$ due to jumps occur at the finite set of points $\{t_1, \ldots, t_n \mid t_i \in [0, T]\}$. Therefore, for every $k > 0$, the set $S_k = \{t \mid \mathcal{I}(t) > k\}$ is finite, leading to the immediate derivation of the set
	\begin{equation}
		S=\bigcup_{n\in\mathbb{N}}S_{\frac{1}{n}}=\{t:\Delta \mathcal{I}\neq 0\},
	\end{equation}
	which is also finite over the bounded interval $[0,T]$. 
	
	The Markov chain $M_t$ takes values in a finite set of integers, so there exists a constant $K$ such that $\Delta < K$. If $T < \infty$, a stopping time sequence $\{\tau_k\}_{k \geq 0}$ can be found such that, for almost every $\omega \in \Omega$, there is a finite random variable $\bar{k} = \bar{k}(\omega)$ satisfying $0 = \tau_0 < \tau_1 < \ldots < \tau_{\bar{k}} = T$, and for $k \geq \bar{k}$, $\tau_k = T$. Thus, on each interval between jumps $[\tau_k, \tau_{k+1})$, for a fixed $\omega \in \Omega$, $M_t$ is a random constant. This implies that the instantaneous equation parameters of this complex system over each random time interval can be regarded as a specific value $\theta$ in the parameter switching space $\Theta$. Consequently, the convergence of DSBL can be reduced to the convergence of SBL on the local time intervals $[\tau_k, \tau_{k+1})$. The proof of SBL's convergence is provided below.
	
	According to Eq.\ref{Appendix_loss}, we define a mapping from a point to a set, formulated as follows
	\begin{align}
		\mathcal{G}: \quad \Omega & \longrightarrow \sigma(\Omega), \\
		(\theta, \gamma) & \longmapsto \arg \min _{(\bar{\theta}, \bar{\gamma}) \in \Omega} \widehat{\mathcal{L}}(\bar{\theta}, \bar{\gamma}; \theta, \gamma),
	\end{align}
	where \(\sigma(\Omega)\) represents the \(\sigma\)-algebra generated by \(\Omega\). Clearly, \(\left(\theta^{(k+1)}, \gamma^{(k+1)}\right) \in \mathcal{G}\left(\theta^{(k)}, \gamma^{(k)}\right)\). Specifically, if the dictionary matrix \(D\) is full column rank, then \(\mathcal{G}(\theta, \gamma)\) is a singleton in \(\Omega\), reducing the point-to-set mapping to a point-to-point mapping. A fixed point of the mapping \(\mathcal{G}\) is a point \((\theta, \gamma)\) that satisfies \(\{(\theta,\gamma)\}=\mathcal{G}(\theta,\gamma)\), while a generalized fixed point of the mapping \(\mathcal{G}\) is a point \((\theta, \gamma)\) that satisfies \((\theta, \gamma) \in \mathcal{G}(\theta, \gamma)\). 
	
	Let $S$ be the set of generalized fixed points of $\mathcal{G}$. By Theorem~\ref{bounded sequence}, we assume $\left\{\left(\theta^{(k)}, \gamma^{(k)}\right)\right\}_{k=0}^{\infty}$ is an iterative sequence generated by the point-to-set mapping $\mathcal{G}$, starting from the initial point $\left(\theta^{(0)}, \gamma^{(0)}\right) \in \operatorname{int}(\Omega)$. Since the sequence $\left\{\theta^{(k)}\right\}_{k=0}^{\infty}$ is bounded, let $m_{\theta}$ and $M_{\theta}$ be the lower and upper bounds of the sequence $\left\{\theta^{(k)}\right\}_{k=0}^{\infty}$, respectively. Therefore, there exists a compact set $C=\left[m_{\theta}, M_{\theta}\right]^{m} \times\left[0, M_{\gamma}\right]^{m} \subseteq \Omega$ such that for any $k=0,1,\ldots$, all points $\left(\theta^{(k)}, \gamma^{(k)}\right)$ are within this compact set $C$. Thus, there exists the following convergent subsequence
	\begin{equation}\label{30}
		\left(\theta^{\left(k_{s}\right)}, \gamma^{\left(k_{s}\right)}\right) \longrightarrow\left(\theta^{*}, \gamma^{*}\right) \text {, as } s \rightarrow \infty.
	\end{equation}
	
	According to the definition of $\mathcal{G}$, for any $(\bar{\theta}, \bar{\gamma}) \in \mathcal{G}(\theta, \gamma)$, we have
	\begin{equation}\label{27}
		\mathcal{L}(\bar{\theta}, \bar{\gamma}) \leq \widehat{\mathcal{L}}(\bar{\theta}, \bar{\gamma} ; \theta, \gamma) \leq \widehat{\mathcal{L}}(\theta, \gamma ; \theta, \gamma)=\mathcal{L}(\theta, \gamma),
	\end{equation}
	i.e., $\mathcal{G}$ is monotonically non-increasing with respect to $\mathcal{L}$. Combining the monotonicity of $\mathcal{G}$ with respect to $\mathcal{L}$, we have
	\begin{equation}
		\mathcal{L}\left(\theta^{\left(k_{s+1}\right)}, \gamma^{\left(k_{s+1}\right)}\right) \leq \mathcal{L}\left(\theta^{\left(k_{s}\right)}, \gamma^{\left(k_{s}\right)}\right)<+\infty.
	\end{equation}
	
	From the previous assumptions, it can be inferred that $\left(\theta, \gamma\right) \in \operatorname{int}(\Omega) \cup \Omega_{2}$. Actually, observing from \ref{30}, if $\left(\theta, \gamma\right) \in \Omega_{1}$ then we have
	\begin{equation}
		\lim _{s \rightarrow \infty} \mathcal{L}\left(\theta^{\left(k_{s}\right)}, \gamma^{\left(k_{s}\right)}\right)=+\infty.
	\end{equation}
	This contradicts \ref{30}.
	
	Next, if $\left(\theta, \gamma\right) \in \operatorname{int}(\Omega)$, by the continuity of $\mathcal{L}$ on $\operatorname{int}(\Omega)$, we have
	\begin{equation}\label{32}
		\lim _{s \rightarrow \infty} \mathcal{L}\left(\theta^{\left(k_{s}\right)}, \gamma^{\left(k_{s}\right)}\right)=\mathcal{L}\left(\theta^{*}, \gamma^{*}\right).
	\end{equation}
	If $\left(\theta^{*}, \gamma^{*}\right) \in \Omega_{2}$, then according to the supplementary definition of $\mathcal{L}$ on $\Omega_{2}$, \ref{32} holds.
	
	On the other hand, for the sequence ${\left(\theta^{\left(k_{s}+1\right)}, \gamma^{\left(k_{s}+1\right)}\right)}_{s=1}^{\infty}$, there exists a convergent subsequence
	\begin{equation}
		\left(\theta^{\left(k_{s_{j}}+1\right)}, \gamma^{\left(k_{s_{j}}+1\right)}\right) \longrightarrow\left(\theta^{* *}, \gamma^{* *}\right) \quad \text {, as } j \rightarrow \infty.
	\end{equation}
	
	Similar to the proof of \ref{32}, it follows that
	\begin{equation}\label{34}
		\lim _{j \rightarrow \infty} \mathcal{L}\left(\theta^{\left(k_{s_{j}}+1\right)}, \gamma^{\left(k_{s_{j}}+1\right)}\right)=\mathcal{L}\left(\theta^{* *}, \gamma^{* *}\right).
	\end{equation}
	
	Since $\mathcal{G}$ is monotonically non-increasing with respect to $\mathcal{L}$, the sequence $\left\{\mathcal{L}\left(\theta^{(k)}, \gamma^{(k)}\right)\right\}_{k=0}^{\infty}$ is monotonically bounded, and hence convergent. Combining \ref{32} and \ref{34}, then
	\begin{equation}\label{35}
		\mathcal{L}\left(\theta^{*}, \gamma^{*}\right)=\mathcal{L}\left(\theta^{* *}, \gamma^{* *}\right).
	\end{equation}
	
	Furthermore, from \ref{30} and \ref{32}, we conclude that
	\begin{equation}
		\left(\theta^{\left(k_{s_{j}}\right)}, \gamma^{\left(k_{s_{j}}\right)}\right) \longrightarrow\left(\theta^{*}, \gamma^{*}\right) \text {, as } j \rightarrow \infty,
	\end{equation}
	\begin{equation}
		\lim _{j \rightarrow \infty} \mathcal{L}\left(\theta^{\left(k_{s_{j}}\right)}, \gamma^{\left(k_{s_{j}}\right)}\right)=\mathcal{L}\left(\theta^{*}, \gamma^{*}\right).
	\end{equation}
	
	Since the two sequences satisfy
	\begin{equation}
		\left(\theta^{\left(k_{s_{j}}+1\right)}, \gamma^{\left(k_{s_{j}}+1\right)}\right) \in \mathcal{G}\left(\theta^{\left(k_{s_{j}}\right)}, \gamma^{\left(k_{s_{j}}\right)}\right),
	\end{equation}
	and both $\left\{\left(\theta^{\left(k_{s_{j}}\right)}, \gamma^{\left(k_{s_{j}}\right)}\right)\right\}$ and $\left\{\left(\bar{\theta}^{\left(k_{s_{j}}+1\right)}, \bar{\gamma}^{\left(k_{s_{j}}+1\right)}\right)\right\}$ are in $\operatorname{int}(\Omega) \cup \Omega_{2}$. Then if $\left\{\bar{\gamma}_{i}^{(k)} \neq 0\right\}$, $\left|\bar{\theta}_{i}^{(k)}\right| / \bar{\gamma}_{i}^{(k)}$ is bounded. $\bar{\theta}_{i}^{(k)}$ and $\bar{\gamma}_{i}^{(k)}$ share the same zero properties.
	
	Thus, based on the definition of $\mathcal{L}$ and the existence of $\lim_{k \rightarrow \infty} \widehat{\mathcal{L}}\left(\bar{\theta}^{(k)}, \bar{\gamma}^{(k)} ; \theta^{(k)}, \gamma^{(k)}\right)$, we can conclude that the sequence $\left\{\widehat{\mathcal{L}}\left(\bar{\theta}^{(k)}, \bar{\gamma}^{(k)} ; \theta^{(k)}, \gamma^{(k)}\right)\right\}_{k=0}^{\infty}$ is bounded.
	
	If $\left(\theta^{**}, \gamma^{**}\right) \in \operatorname{int}(\Omega)$, then by the continuity of $\widehat{\mathcal{L}}\left(\theta, \gamma ; \theta^*, \gamma^*\right)$ on $\operatorname{int}(\Omega)$, the following holds
	\begin{equation}\label{62}
		\lim _{k \rightarrow \infty} \widehat{\mathcal{L}}\left(\bar{\theta}^{(k)}, \bar{\gamma}^{(k)} ; \theta^*, \gamma^*\right)=\widehat{\mathcal{L}}\left(\theta^{**}, \gamma^{**} ; \theta^*, \gamma^*\right).
	\end{equation}
	If $\left(\theta^{**}, \gamma^{**}\right) \in \Omega_2$, then by the definition of $\widehat{\mathcal{L}}\left(\theta, \gamma ; \theta^*, \gamma^*\right)$ on $\Omega_2$, the above equation also holds.
	
	Furthermore, we obtain
	\begin{equation}\label{61}
		\lim _{k \rightarrow \infty} \widehat{\mathcal{L}}\left(\bar{\theta}^{(k)}, \bar{\gamma}^{(k)} ; \theta^{(k)}, \gamma^{(k)}\right)=\widehat{\mathcal{L}}\left(\theta^{**}, \gamma^{**} ; \theta^*, \gamma^*\right).
	\end{equation}
	
	Using proof by contradiction, assume $\left(\theta^{**}, \gamma^{**}\right) \notin \mathcal{G}\left(\theta^*, \gamma^*\right)$. Given that $\widehat{\mathcal{L}}\left(\theta, \gamma ; \theta^*, \gamma^*\right)$ is a convex function and $\mathcal{G}\left(\theta^*, \gamma^*\right)$ is non-empty, let $(\tilde{\theta}, \tilde{\gamma})$ be any point in $\mathcal{G}\left(\theta^*, \gamma^*\right)$. Therefore, 
	\begin{equation}
		\widehat{\mathcal{L}}\left(\tilde{\theta}, \tilde{\gamma} ; \theta^*, \gamma^*\right)<\widehat{\mathcal{L}}\left(\theta^{**}, \gamma^{**} ; \theta^*, \gamma^*\right).
	\end{equation}
	Choose a sufficiently small positive number $\varepsilon_1$ such that
	\begin{equation}
		\widehat{\mathcal{L}}\left(\theta^{**}, \gamma^{**} ; \theta^*, \gamma^*\right)-\widehat{\mathcal{L}}\left(\tilde{\theta}, \tilde{\gamma} ; \theta^*, \gamma^*\right) > 2\varepsilon_1.
	\end{equation}
	According to equation \ref{61}, there exists a positive number $k_1$ such that for any $k>k_1$
	\begin{equation}
		\widehat{\mathcal{L}}\left(\bar{\theta}^{(k)}, \bar{\gamma}^{(k)} ; \theta^{(k)}, \gamma^{(k)}\right)>\widehat{\mathcal{L}}\left(\theta^{**}, \gamma^{**} ; \theta^*, \gamma^*\right)-\varepsilon_1.
	\end{equation}
	Given the definition of $\widehat{\mathcal{L}}(\tilde{\theta}, \tilde{\gamma} ; \theta, \gamma)$, we know that for $(\theta, \gamma)$, $\widehat{\mathcal{L}}(\tilde{\theta}, \tilde{\gamma} ; \theta, \gamma)$ is continuous on $\Omega$. Therefore, there exists a positive number $k_2$ such that for any $k>k_2$
	\begin{equation}
		\widehat{\mathcal{L}}\left(\tilde{\theta}, \tilde{\gamma} ; \theta^{(k)}, \gamma^{(k)}\right)<\widehat{\mathcal{L}}\left(\tilde{\theta}, \tilde{\gamma} ; \theta^*, \gamma^*\right)+\varepsilon_1.
	\end{equation}
	Combining the above parts, we can deduce that for any $k>\max\left\{k_1, k_2\right\}$, the following holds
	\begin{equation}\label{123}
		\widehat{\mathcal{L}}\left(\bar{\theta}^{(k)}, \bar{\gamma}^{(k)} ; \theta^{(k)}, \gamma^{(k)}\right)>\widehat{\mathcal{L}}\left(\tilde{\theta}, \tilde{\gamma} ; \theta^{(k)}, \gamma^{(k)}\right).
	\end{equation}
	
	However, since
	\begin{equation}
		\left(\bar{\theta}^{(k)}, \bar{\gamma}^{(k)}\right) \in \mathcal{G}\left(\theta^{(k)}, \gamma^{(k)}\right).
	\end{equation}
	We have
	\begin{equation}
		\widehat{\mathcal{L}}\left(\bar{\theta}^{(k)}, \bar{\gamma}^{(k)} ; \theta^{(k)}, \gamma^{(k)}\right) \leq \widehat{\mathcal{L}}\left(\theta, \gamma ; \theta^{(k)}, \gamma^{(k)}\right),
	\end{equation}
	holds for any $(\theta, \gamma) \in \Omega$, leading to a contradiction with equation \ref{123}. Thus, we have
	\begin{equation}\label{38}
		\left(\theta^{**}, \gamma^{**}\right) \in \mathcal{G}\left(\theta^{*}, \gamma^{*}\right).
	\end{equation}
	
	By equation \ref{27}, $\mathcal{G}$ is monotonically non-increasing with respect to $\mathcal{L}$. If $(\theta, \gamma) \notin S$, then $\forall (\bar{\theta}, \bar{\gamma}) \in \mathcal{G}$, $\mathcal{L}(\bar{\theta}, \bar{\gamma}) < \mathcal{L}(\theta, \gamma)$. In fact, if $\mathcal{L}(\bar{\theta}, \bar{\gamma}) = \mathcal{L}(\theta, \gamma)$, by equation \ref{27}, we know that $\widehat{\mathcal{L}}(\bar{\theta}, \bar{\gamma} ; \theta, \gamma) = \widehat{\mathcal{L}}(\theta, \gamma ; \theta, \gamma)$, and thus $(\theta, \gamma) \in S$. From equations \ref{35} and \ref{38}, we have
	\begin{equation}
		\left(\theta^{*}, \gamma^{*}\right) \in S.
	\end{equation}
	
	According to the iteration generated by the mapping $\mathcal{G}$ in equation \ref{Appendix_loss}, there exists $z^{*} \in \partial\left|\theta^{*}\right|_{1}$ such that
	\begin{equation}\label{26}
		\left\{\begin{array}{l}
			D_{i}^{T}\left(D \theta^{*}-y\right)+\sigma^{2} \sqrt{c_{i}^{*}} z_{i}^{*}=0, \\
			\left|\theta_{i}^{*}\right| / \sqrt{c_{i}^{*}}=\gamma_{i}^{*}, i=1, \cdots, m,\\
			c^{*}:=\nabla_{\gamma} \log \left|\sigma^{2} I_{n}+D \Gamma D^{T}\right| \Big |_{\gamma=\gamma^{*}}.
		\end{array}\right.
	\end{equation}
	
	It can be seen that Eq.\ref{26} represents the KKT conditions of Eq.\ref{loss_L}. Therefore, $\left(\theta^{*}, \gamma^{*}\right)$ is also a stationary point of Eq.\ref{loss_L}. Finally, by the convergence of $\left\{\mathcal{L}\left(\theta^{(k)}, \gamma^{(k)}\right)\right\}_{k=0}^{\infty}$ and Eq.\ref{32}, it follows that the iterative sequence $\left\{\left(\theta^{(k)}, \gamma^{(k)}\right)\right\}_{k=0}^{\infty}$, generated by the mapping $\mathcal{G}$, converges to a stationary point $\left(\theta^{*}, \gamma^{*}\right)$ of Eq.\ref{loss_L}, i.e.,
	\begin{equation}\label{29}
		\lim _{k \rightarrow \infty} \mathcal{L}\left(\theta^{(k)}, \gamma^{(k)}\right)=\mathcal{L}\left(\theta^{*}, \gamma^{*}\right).
	\end{equation}
	
	This completes the proof of the theorem.
	
	\subsection{A proof of Theorem~\ref{Error bound}}\label{Appendix2}
	$Proof:$ For the iterative sequence \(\left\{\left(\theta^{(k)}, \gamma^{(k)}\right)\right\}_{k=0}^{\infty}\) generated by the algorithm, one could verify that
	\begin{align}\label{41}
		\left\{
		\begin{array}{l}
			D_{i}^{T}\left(D \theta^{(k+1)}-y\right)+\sigma^{2} \sqrt{c_{i}^{(k)}} z_{i}^{(k+1)}=0, \\
			\left|\theta_{i}^{(k+1)}\right| / \sqrt{c_{i}^{(k)}}=\gamma_{i}^{(k+1)}, \: i=1, \cdots, m,
		\end{array}
		\right.
	\end{align}
	where
	\begin{align} c^{(k)}&=\operatorname{diag}\left[D^{T}\left(\sigma^{2} I_{n}+D \Gamma^{(k)} D^{T}\right)^{-1} D\right],\\
		z^{(k+1)}&=\left[z_{1}^{(k+1)}, \cdots, z_{n}^{(k+1)}\right]^{T} \in \partial\left\|\theta^{(k+1)}\right\|_{1}.
	\end{align}

	We define a truncation set
	\begin{equation}
		\mathbb{M}(\theta) := \{i : |\theta_i| > \tau\},
	\end{equation}
	where \(\tau\) is a fixed threshold. When \(\tau < \theta_{\text{min}}\), we have
	\begin{equation}
		\theta_{\text{min}} = \min_{i \in \mathbb{M}(\theta^{\text{true}})} |\theta_i^{\text{true}}|.
	\end{equation}
	
	Similar to Theorem \ref{convergence}, it is advantageous to consider error bound over the local time interval $[\tau_k,\tau_{k+1})$. For \( D^{T} D = I_{m} \), we have
	\begin{align}
		\left(\sigma^{2} I_{n} + D \Gamma^{(k+1)} D^{T}\right)^{-1} = \sigma^{-2} I_{n} - \sigma^{-2} D \Gamma^{(k+1)} \left[ \sigma^{2} I_{m} + \Gamma^{(k+1)} \right]^{-1} D^{T},
	\end{align}
	and
	\begin{align}\label{42}
		c_{i}^{(k+1)} 
		&= \sigma^{-2} D_{i}^{T} D_{i} - \sigma^{-2} D_{i}^{T} D \Gamma^{(k+1)} \left[\sigma^{2} I_{m} + \Gamma^{(k+1)}\right]^{-1} D^{T} D_{i} \notag \\
		&= \sigma^{-2} - \sigma^{-2} e_{i}^{T} \Gamma^{(k+1)} \left[\sigma^{2} I_{m} + \Gamma^{(k+1)}\right]^{-1} e_{i} \notag\\
		&= \sigma^{-2} - \sigma^{-2} \frac{\gamma_{i}^{(k+1)}}{\sigma^{2} + \gamma_{i}^{(k+1)}} \notag \\
		&= \frac{\sqrt{c_{i}^{(k)}}}{\sigma^{2} \sqrt{c_{i}^{(k)}} + \left|\theta_{i}^{(k+1)}\right|}.
	\end{align}
	
	According to Theorem~\ref{convergence}, the sequence \(\left\{\left(\theta^{(k)}, \gamma^{(k)}\right)\right\}\) converges to the stationary point \(\left(\theta^{*}, \gamma^{*}\right)\) of \ref{loss_L}. All stationary points of \ref{loss_L} satisfy the KKT conditions \ref{26}, hence \(c^{(k)}\) converges to \(c^{*}\) as $k\rightarrow \infty$. Taking the limit of \ref{42} yields an implicit equation
	\begin{equation}
		\sigma^{2} c_{i}^{*} + \left|\theta_{i}^{*}\right| \sqrt{c_{i}^{*}} - 1 = 0.
	\end{equation}
	Solving this quadratic equation for \(\sqrt{c_i^{*}}\) gives
	\begin{align}\label{44}
		\sqrt{c_{i}^{*}} = \frac{-\left|\theta_{i}^{*}\right| + \sqrt{\left|\theta_{i}^{*}\right|^{2} + 4 \sigma^{2}}}{2 \sigma^{2}} = \frac{2}{\left|\theta_{i}^{*}\right| + \sqrt{\left|\theta_{i}^{*}\right|^{2} + 4 \sigma^{2}}}.
	\end{align}
	
	From \ref{41}, we obtain
	\begin{align}\label{45}
		D_{i}^{T}\left(D \theta^{*} - y\right) + \sigma^{2} \sqrt{c_{i}^{*}} z_{i}^{*} = 0,
	\end{align}
	where \(z^{*} := \left[z_{1}^{*}, z_{2}^{*}, \cdots, z_{m}^{*}\right]^{T}\). Substituting \ref{Regression} and \ref{44} into \ref{45}, we get
	\begin{align}\label{46}
		\theta_{i}^{*} - \theta_{i}^{\text {true}} + \frac{-\left|\theta_{i}^{*}\right| + \sqrt{\left|\theta_{i}^{*}\right|^{2} + 4 \sigma^{2}}}{2} z_{i}^{*} = D_{i}^{T} \varepsilon.
	\end{align}
	
	For any \(i \in \mathbb{M}\left(\theta^{\text {true}}\right)\), if \(\left|D_{i}^{T} \varepsilon\right| < \frac{1}{2} \theta_{\min}\), then \(i \in \mathbb{M}\left(\theta^{*}\right)\). Indeed, if \(i \notin \mathbb{M}\left(\theta^{*}\right)\), then from \ref{46}, we have
	\begin{align}\label{47}
		\theta_{i}^{*} + \frac{-\left|\theta_{i}^{*}\right| + \sqrt{\left|\theta_{i}^{*}\right|^{2} + 4 \sigma^{2}}}{2} z_{i}^{*} = \theta_{i}^{\text {true}} + D_{i}^{T} \varepsilon.
	\end{align}
	
	When \(\left|D_{i}^{T} \varepsilon\right| < \frac{1}{2} \theta_{\min}\), there exists
	\begin{equation}
		\left|\theta_{i}^{\text {true}} + D_{i}^{T} \varepsilon\right| \geq \left|\theta_{i}^{\text {true}}\right| - \left|D_{i}^{T} \varepsilon\right| \geq \frac{1}{2} \theta_{\min},
	\end{equation}
	and
	\begin{equation}
		\left|\theta_{i}^{*} + \frac{-\left|\theta_{i}^{*}\right| + \sqrt{\left|\theta_{i}^{*}\right|^{2} + 4 \sigma^{2}}}{2} z_{i}^{*}\right| \leq \tau + \sigma.
	\end{equation}
	
	By assumption, this contradicts \ref{47}. For any \(i \notin \mathbb{M}\left(\theta^{\text {true}}\right)\), if \(\left|D_{i}^{T} \varepsilon\right| < \tau\), then \(i \notin \mathbb{M}\left(\theta^{*}\right)\). Indeed, if \(i \in \mathbb{M}\left(\theta^{*}\right)\), then from \ref{46} and the assumption, we have
	\begin{equation}
		\left|D_{i}^{T} \varepsilon\right| = \frac{\left|\theta_{i}^{*} + \sqrt{\left|\theta_{i}^{*}\right|^{2} + 4 \sigma^{2}} z_{i}^{*}\right|}{2} > \tau,
	\end{equation}
	which contradicts the assumption. Thus, on \(\left\{\max _{i}\left|D_{i}^{T} \varepsilon\right| < \tau\right\}\), we have \(\mathbb{M}\left(\theta^{\text {true}}\right) = \mathbb{M}\left(\theta^{*}\right)\). Given \(D^{T} D = I_{m}\), the values \(D_{i}^{T} \varepsilon, i = 1, \cdots, m\) are independent Gaussian random variables with mean zero and variance \(\sigma^{2}\). By Hoeffding's inequality, we have
	\begin{equation}
		\begin{aligned}
			& \mathbb{P}\left[\mathbb{M}\left(\theta^{*}\right) = \mathbb{M}\left(\theta^{\text{true}}\right)\right] =  \mathbb{P}\left[\max _{i}\left|D_{i}^{T} \varepsilon\right| < \tau\right]  \\ = & \prod_{i} \mathbb{P}\left[\left|D_{i}^{T} \varepsilon\right| < \tau\right] \geq  \left(1 - 2 e^{-\tau^{2} / 2 \sigma^{2}}\right)^{m}  \geq 1 - e^{-\tau^{2} / 4 \sigma^{2}}.
		\end{aligned}
	\end{equation}
	
	For any \(i \in \mathbb{M}\left(\theta^{\text {true}}\right)\), by Taylor's theorem, there exists \(\eta \in (0, 4 \sigma^{2})\) such that
	\begin{equation}
		\sqrt{\left|\theta_{i}^{*}\right|^{2} + 4 \sigma^{2}} = \left|\theta_{i}^{*}\right| + \frac{2 \sigma^{2}}{\sqrt{\left|\theta_{i}^{*}\right|^{2} + \eta}}.
	\end{equation}
	
	Substituting this into \ref{46} gives
	\begin{equation}
		\left|\theta_{i}^{\theta_{true}} - \theta_{i}^{*}\right| = \left|D_{i}^{T} \varepsilon - \frac{\sigma^{2}}{\sqrt{\left|\theta_{i}^{*}\right|^{2} + \eta}}\right| \leq \tau + \frac{\sigma^{2}}{\tau}.
	\end{equation}
	
	This yields the algorithmic error estimate for the instantaneous parameter \(\theta\) over the local time interval \([\tau_k, \tau_{k+1})\). On the entire time interval \([0, T]\), there is the unknown parameter vector \(\vartheta\) of the equation as \(d\)-dimensional, and the Markov chain \(M_t\) containing a finite number of \(K\) states. The error across the entire interval $[0,T]$ can be recursively determined.
	\begin{equation}\label{error bound appendix}
		\left|\theta^{true} - \theta^{*}\right| \leq \sqrt{dK} \tau + \frac{\sqrt{dK} \sigma^{2}}{\tau}.
	\end{equation}
	
	Finally, under the condition of Theorem~\ref{Error bound}, where \(2\sigma\sqrt{\log 2m} \leq \tau < \frac{1}{2}\theta_{\min} - \sigma\), the uniform error bound is guaranteed. Observe that the RHS of \ref{error bound appendix} takes the form \( x + \frac{a}{x} \), where \( a > 0 \), indicating that \(\tau\) attains its minimum within the first quadrant when \(\tau = \sigma\). Assuming \( 2\sigma\sqrt{\log 2m} < \sigma \), it follows that \( m \leq \frac{1}{2}e^{1/4} \approx 0.64 \), which contradicts the fact that \( m \) is a positive integer. Therefore, \( \sigma < 2\sigma\sqrt{\log 2m} < \frac{1}{2}\theta_{\min} - \sigma \). When the positive value \(\tau\) is set to \(\frac{1}{2}\theta_{\min} - \sigma\), there exists a consistent upper bound for the algorithm error. Substituting this into the error bound we yield that
	\begin{equation}
		|\theta^{\text{true}} - \theta^*|
		\leq \sqrt{dK} \frac{\theta_{\min}^2 - 4\sigma\theta_{\min} + 8\sigma^2}{2\theta_{\min} - 4\sigma}.
	\end{equation} 
	
	This completes the proof of the theorem.

	\bibliographystyle{elsarticle-num} 
	\bibliography{references} 
\end{document}